\documentclass[lettersize,journal]{IEEEtran}
\usepackage{amsmath,amsfonts,amssymb}
\usepackage{mathtools}
\usepackage{algorithmic}
\usepackage{algorithm}
\usepackage{array}
\usepackage[caption=false,font=normalsize,labelfont=sf,textfont=sf]{subfig}
\usepackage{textcomp}
\usepackage{stfloats}
\usepackage{url}
\usepackage{verbatim}
\usepackage{graphicx}
\usepackage[table,dvipsnames]{xcolor}
\usepackage{cite}
\usepackage{hyperref}
\hypersetup{
    colorlinks=true,
    linkcolor=blue,
    filecolor=blue,      
    urlcolor=blue,
    citecolor=blue,
    }
\usepackage{booktabs}
\usepackage{multirow}
\usepackage{rotating}  
\usepackage{subcaption}
\usepackage{subfig}
\usepackage{amsthm}
\usepackage{caption}
\usepackage{footnotehyper}

\newcommand{\R}{\mathbb R}

\newcommand{\bmx}[1]{\begin{bmatrix}#1\end{bmatrix}} 
\newcommand{\pth}[1]{\left(#1\right)} 

\newcommand{\pdrv}[2]{\frac{\partial #1}{\partial #2}}

\DeclarePairedDelimiter{\ceil}{\lceil}{\rceil}
\DeclarePairedDelimiter{\floor}{\lfloor}{\rfloor}
\DeclarePairedDelimiter{\abs}{\lvert}{\rvert}

\newcommand{\rarr}{\rightarrow} 

\makeatletter
\let\oldceil\ceil
\def\ceil{\@ifstar{\oldceil}{\oldceil*}}
\makeatother
\makeatletter
\let\oldfloor\floor
\def\floor{\@ifstar{\oldfloor}{\oldfloor*}}
\makeatother
\makeatletter
\let\oldnorm\norm
\def\norm{\@ifstar{\oldnorm}{\oldnorm*}}
\makeatother
\makeatletter
\let\oldabs\abs
\def\abs{\@ifstar{\oldabs}{\oldabs*}}
\makeatother

\begin{document}

\title{Automatic Grid Updates for Kolmogorov–Arnold Networks using Layer Histograms}

\author{Jamison Moody and James Usevitch
\thanks{Jamison Moody and James Usevitch are with the Department of Electrical and Computer Engineering, Brigham Young University, Provo, UT 84602. {\tt\small \{jmm1995,james\_usevitch\}@byu.edu}}
}

\markboth{}%
{Shell \MakeLowercase{\textit{et al.}}: A Sample Article Using IEEEtran.cls for IEEE Journals}


\maketitle

\begin{abstract}
Kolmogorov-Arnold Networks (KANs) are a class of neural networks that have received increased attention in recent literature.
In contrast to MLPs, KANs leverage parameterized, trainable activation functions and offer several benefits including improved interpretability and higher accuracy on learning symbolic equations. However, 
the original KAN architecture requires adjustments to the domain discretization of the network (called the ``domain grid") during training, creating extra overhead for the user in the training process. Typical KAN layers are not designed with the ability to autonomously update their domains in a data-driven manner informed by the changing output ranges of previous layers. 
To address this limitation, we propose a new variant of KANs called AdaptKAN that automatically adapts the domain grid during the training process.
This is accomplished by leveraging a novel histogram algorithm that tracks layer inputs both within and outside the current domain grid using an exponential moving average of the input distribution.
As an added benefit, this histogram algorithm may also be applied towards detecting out-of-distribution (OOD) inputs in a variety of settings. 
We demonstrate that AdaptKAN exceeds or matches the performance of prior KAN architectures and MLPs on four different tasks: learning scientific equations from the Feynman dataset, image classification from frozen features, learning a control Lyapunov function, and detecting OOD inputs on the OpenOOD v1.5 benchmark.
\end{abstract}

\begin{figure*}[!t]
\centering
\includegraphics[width=\textwidth]{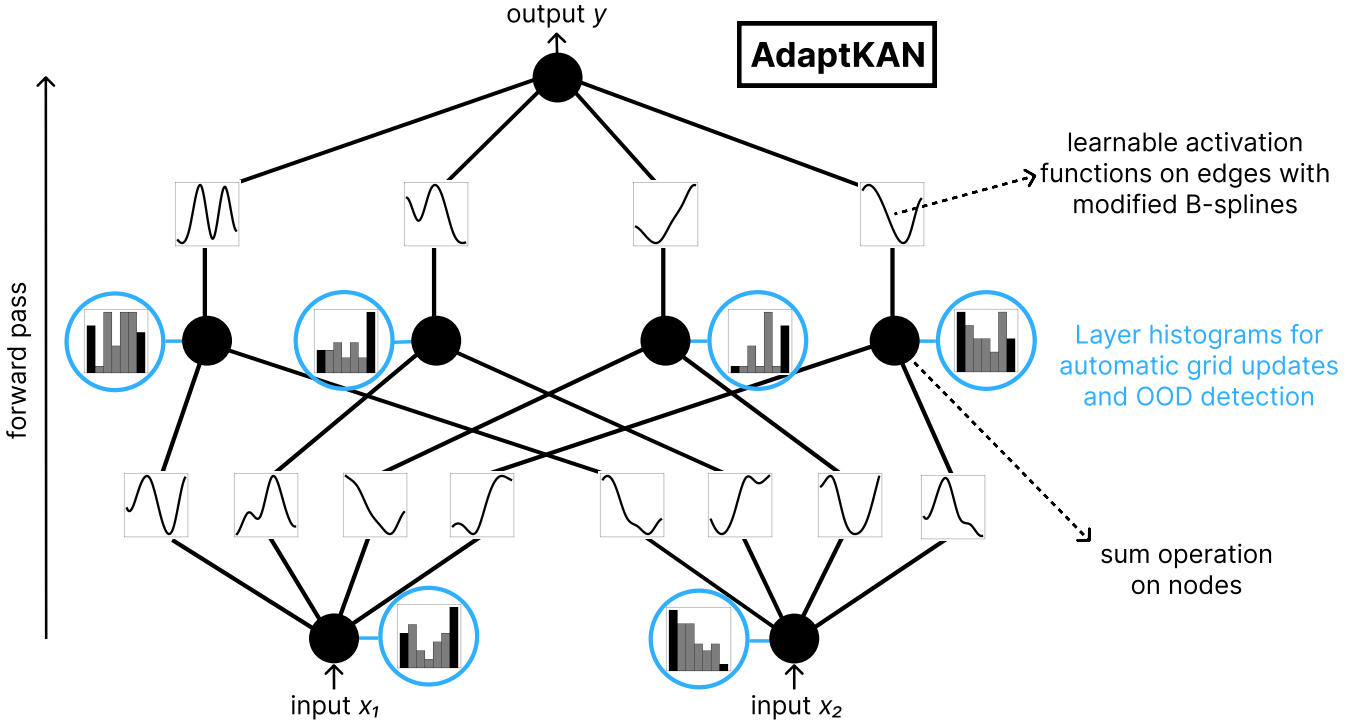}
\caption{A high level overview of our proposed AdaptKAN architecture. AdaptKAN augments the original KAN architecture with histograms that approximate marginal distributions for each input feature to each layer. The number of bins is equal to the number of grid intervals for each B-spline. AdaptKAN stores two extra bins per histogram (depicted as solid black bins in the figure) that keep track of how much data is falling outside of the current grid domain. If the out-of-domain bin counts rise above a specified threshold, AdaptKAN stretches the domain. On the other hand, if the edge histogram bin counts fall below a specified threshold, AdaptKAN shrinks the domain. Further details on this adaptation algorithm are provided in Section \ref{sec:main_results}.}
\label{fig:main_diagram}
\end{figure*}


\section{Introduction}
\IEEEPARstart{D}{eep} Learning has emerged as a promising tool for scientific discovery with its ability to uncover laws underlying physical systems \cite{liu2024kan, kim2020integration, yu2024learning} and predict complex biological phenomena \cite{abramson2024accurate, lotfollahi2019scgen, aalto2020gene}. 
One promising class of architectures that has shown favorable performance in scientific domains is that of Kolmogorov–Arnold Networks (KANs), which were introduced in \cite{liu2024kan}. These networks 
use trainable B-spline activation functions that facilitate translation of the network back into closed-form scientific formulae. 
Since the original KAN paper was introduced, there have been a myriad of variants to the original architecture proposed including activation function alternatives \cite{seydi2024unveiling, ss2024chebyshev, aghaei2025fkan, ta2024bsrbf, li2024kolmogorov} and applications to specific architectures such as CNNs \cite{bodner2024convolutional}, GNNs \cite{zhang2406graphkan}, Transformers \cite{yang2024kolmogorov}, and U-Nets \cite{li2025u}.

Unlike MLPs, KANs typically have a strict domain defined for each of their parameterized activation functions in each layer. This is represented by a grid on some interval $[a,b]$ split into a certain number of heterogeneous intervals. 
One limitation of KANs is that this domain grid $[a,b]$ and the heterogeneous intervals on the domain need to be adapted during training in response to the varying range of the outputs from previous layers. The authors of \cite{liu2024kan} deal with this issue by setting multiple hyperparameters relating starting, stopping, and when to adapt the domain grid. The grid is altered based on a single batch of data, ignoring past data distribution history. This heuristic can cause issues in scenarios where there is a high sensitivity to incoming data. This may include scenarios where the data contains outlier or ``poisoned" examples. Such scenarios arise in real-world settings such as 
spam filtering, offensive chatbots, corrupted deep reinforcement learning inputs, and sensitivity to boundary data for Physics Informed Neural Networks \cite{cina2024machine, kiourti2020trojdrl, bajaj2023recipes}. 

One way to deal with outlier detection in safety critical applications is out-of-distribution (OOD) detection. Many of these methods require a designing a specific training task geared towards detection \cite{papadopoulos2021outlier, winkens2020contrastive, hendrycks2019using} and/or exposure to outliers during the training process \cite{NEURIPS2024_ee204617, hendrycks2018deep, du2022vos}. 
In the area of OOD detection for image classification, the OpenOOD v1.5 evaluation benchmark \cite{zhang2023openood} is well-respected and provides a variety of reproducible standardized evaluations including Near-OOD (OOD examples that are semantically similar to the original training data) and Far-OOD (OOD examples that have a larger semantic difference compared to in-distribution samples).
Many of the requirements for prior OOD algorithms can be difficult to achieve in practice. As an alternative, post-hoc OOD detection methods perform detection without altering the original training objective or requiring OOD samples. This is useful in real-world application scenarios where retraining the network is difficult \cite{yang2024generalized}. 
Finding accurate and effective post-hoc OOD algorithms for neural network architectures remains an important open problem.

In an effort to address the issues of automatic grid adjustments and OOD detection for KANs, we present a new architecture called \emph{AdaptKAN}.
This architecture contains a novel algorithm that leverages exponential moving-average histograms to automatically adjust the domain grid for each of its layers in response to incoming data. Our histogram approach also exhibits the additional benefit of acting as an OOD detector that achieves state-of-the-art performance on the OpenOOD 1.5 benchmark.
More specifically, our contributions are as follows:

\begin{enumerate}
    \item We present a novel KAN architecture (AdaptKAN) that leverages a histogram method to adapt grid domains based on incoming data. Our method eliminates the need to select the timing for domain grid adjustment, simplifying and streamlining the KAN training process.
    \item We introduce a novel OOD detection method based on the histogram architecture used by AdaptKAN. We demonstrate that this method achieves state-of-the-art results on the OpenOOD v1.5 benchmark.
    \item We demonstrate the efficacy of AdaptKAN on three tasks: symbolic equation learning, image classification from frozen features, and learning control Lyapunov functions. Our results demonstrate that the AdaptKAN architecture matches or exceeds the performance of prior architectures.
\end{enumerate}

The organization of this paper is as follows. Section \ref{sec:related_works} gives a more thorough overview of related work. Section \ref{sec:main_results} presents our main results, including our novel contributions of the AdaptKAN architecture and histogram-based OOD detection. Section \ref{sec:experimental_results} presents our experimental results demonstrating the efficacy of the AdaptKAN architecture and the histogram-based OOD detection algorithm. A brief conclusion is given in Section \ref{sec:conclusion}.

\section{Background and Related Works}
\label{sec:related_works}

\subsection{Using Data Statistics to improve Neural Network Training}

In order to improve neural network training, data statistics are often utilized during the training process. Batch normalization \cite{ioffe2015batch} uses batch means and variances of features to normalize inputs as they come into the network, which has a variety of benefits including network regularization, the ability to use higher learning rates, and robustness to a variety of initializations. One of the main benefits of batch normalization is its ability to smooth the optimization landscape \cite{santurkar2018does}. Similar to our method, batch normalization uses a running average of population statistics, however for a different purpose: to normalize inputs during inference. Other popular methods that utilize data statistics during training include Layer Normalization \cite{ba2016layer} (useful to stabilize training in recurrent neural networks and does not depend bilon batch size) and Group Normalization \cite{wu2018group} (effective in computer vision tasks where it normalizes across grouped channels). 

Histograms are 
an alternative
way to keep track of network and distribution statistical information corresponding to discrete distributions. For example, \cite{guicquero2022histogram} uses histograms of weight values to help with weight quantization. However, this prior method applies histograms to weight values as opposed to our proposed method of applying histograms to input and layer data distributions. Other prior methods focus on incorporating histograms directly into the backwards pass of the network. For instance, the authors in \cite{ustinova2016learning} introduce a histogram-based loss used when learning deep embeddings. In \cite{sadeghi2022histnet}, histograms of feature maps are calculated from CNNs to improve accuracy in the area of facial expression recognition. Additionally, \cite{sedighi2017histogram} uses a histogram layer to create a statistical distribution of noise residuals from a CNN with the goal to reveal hidden messages in images. The authors in \cite{peeples2021histogram} expand upon this and create their own version of histogram layers for specific applications to texture analysis. These methods analyze histograms of a different architecture (CNNs) compared to our method and require gradient flow through the histograms. In contrast, our proposed method eliminates the need for gradient flow through the histograms. 

\subsection{Domain Adaptation in KANs}

In the original KAN paper \cite{liu2024kan}, specific hyperparameters control when to adjust the domain grid of the network and the adjustment of the grid is based off of a single batch of data. The authors of \cite{liu2024kan} also introduce the ability to do a grid refinement, which creates a ``finer" version of the domain grid and then refits the weights. Some methods avoid the domain grid altogether with B-spline alternatives. Many of these methods involve learnable parameters to auto-adjust the scale or location of the activation functions in the domain \cite{qiu2024relu, li2024kolmogorov, bozorgasl2024wavelet}. However, these methods have some limitations. For example, if activation functions do not have infinite support, data in the region of zero support risks being ignored altogether during training. Other methods such as Wavlet \cite{bozorgasl2024wavelet} and Radial Basis Function (RBF) KANs \cite{ta2024bsrbf} have infinite support, but have a variety of other issues such as a sensitivity to the initial choice of wavelet functions, problematic artifacts introduced by edge effects and numerical instability introduced by large kernel widths \cite{benoudjit2002width}. Data normalization \cite{ss2024chebyshev, qiu2024relu, li2024kolmogorov, bozorgasl2024wavelet} (including batch normalization) is often used, however this can introduce a variety of issues outlined in \cite{aghaei2025fkan}. Bounded activation functions are used as an alternative \cite{aghaei2025fkan}, however bounded activation functions (like the sigmoid function) may be prone to the vanishing gradient problem \cite{gustineli2022survey}.

The authors of \cite{rigas2024adaptive} argue that removing grid adaptation leads to slower convergence and potentially less accurate solutions in the training of Physics Informed KANs. As a result, they introduce a new adaptive grid framework for KANs. Their method is adaptive in the sense that it creates an adaptive optimizer state transition to avoid loss spikes after grid refinement. In contrast to our method, it does not focus on deciding \textit{when} the grid adaptations should occur. Instead, grid adapting is performed at periodic intervals. Additionally, their grid adaptation is based on a single batch of inputs which we demonstrate is more susceptible to data poisoning in Section \ref{results:dist_drift}. The AdaptKAN architecture proposed in this paper instead uses a rolling average of batches for grid adaptation that demonstrates greater robustness to data poisoning.

\subsection{Post-hoc Out-of-Distribution Detection}

OOD Detection is relevant in real-world applications where a network's sensitivity to outlier data could have catastrophic consequences. Many post-hoc OOD detection methods analyze the output of the trained network \cite{liang2017enhancing, liu2020energy, hendrycks2016baseline} or analyze features at different layers of the network \cite{lee2018simple, sastry2020detecting, dong2022neural}.

The work \cite{canevaro2025advancing} uses KANs for Post-hoc OOD detection by comparing the activations of an untrained KAN and multiple trained KANs (trained on a concatenation of CNN network features). However, this involves the storage of $O(N_F N_{C} G)$ KAN model parameters, which can be impractical in real-world scenarios where memory is limited. Here $N_F$ is the number of CNN features, $N_C$ is the number of classes/clusters, and $G$ is the number of intervals in the grid domain. In comparison to this prior approach, our method only involves storing $O(N_F N_B)$ extra parameters. Here $N_B$ is the number of bins for each feature histogram (which can be set to be equal to the number of grid intervals $G$ when used with AdaptKAN). In addition, our method of OOD detection is agnostic to network architecture; however, it is convenient to use with the AdaptKAN architecture because layer histograms are provided as a built-in part of the model.

\subsection{Overview of Kolmogorov-Arnold Networks}
\label{sec:kan_overview}

To provide a better context for our contributions, this section gives an overview of the mathematical formulation of a KAN. Prior work has used the recursive Cox-de Boor Recursion formula for the B-splines \cite{liu2024kan, rigas2024adaptive}. However, we use a non-recursive ``altered" B-spline formula that facilitates indexing into histogram bins and network weights in a straightforward manner. We will first walk through the general definition of a KAN and then outline our unique B-spline formulation.

We denote a KAN layer to be:

\begin{align}
\boldsymbol{\varphi}(\mathbf{x}) &= 
\bmx{
\varphi_{1,1}(\mathbf{x}_1) + \varphi_{1,2}(\mathbf{x}_2) +\dots+ \varphi_{1,n}(\mathbf{x}_{n}) \\
\varphi_{2,1}(\mathbf{x}_1) + \varphi_{2,2}(\mathbf{x}_2) + \dots + \varphi_{2,n}(\mathbf{x}_n) \\
\vdots \\
\varphi_{m,1}(\mathbf{x}_1) +\varphi_{m,2}(\mathbf{x}_2) + \dots + \varphi_{m,n}(\mathbf{x}_n)}
\label{equation:kan_layer_matrix}
\end{align}

Here $n$ and $m$ are the number of inputs and outputs for the KAN layer respectively, and $\mathbf{x} \in \mathbb{R}^n$ are the inputs from the previous layer. We feed each element of $\mathbf{x}$ through different learned univariate activation functions and then sum them up.

These activation functions can be organized into rows and columns based on the indices. For the KAN layer under consideration and $j$-th column of activation functions, we define $\omega_j$ as the number of intervals in the grid domain and indexing function \(\pi : \mathbb{R} \times \mathbb{Z} \rarr \mathbb{Z}\) as
\begin{align}
    \pi(z,j) = \min \left(\floor{\frac{z - a_{j}}{d_{j}}}, \omega_j-1 \right), \label{eq:pi}
\end{align}
where $d_{j} = \frac{b_{j} -a_{j} }{\omega_j}$ is the width of each grid interval on our domain \([a_{j},b_{j}] \subset \R\) and \(a_{j} < b_{j}\). 
This function maps an input \(z\) to an index corresponding to a sub-interval in our domain. 

The interpolation value function \(\theta_j : \R \rarr [0,1]\) for the $j$-th column is defined as:
\begin{align}
    \theta_j(z) = \left\{ \frac{z-a_{j}}{d_{j}} \right\} = \frac{z-a_{j}}{d_{j}} - \floor{\frac{z-a_{j}}{d_{j}}} \label{eq:alpha},
\end{align}

Each activation function in a given KAN layer has a corresponding row vector of weights $\mathbf{w}_{i,j} \in \R^{\omega_j + k}$. We say that $\mathbf{w}_{i,j,\pi(z,j)+s}$ denotes the element of $\mathbf{w}_{i,j}$ at the $\pi(z,j)+s$ index.

The activation function $\varphi_{ij}$ at the $i$-th row and $j$-th column is a B-spline function of degree $k$ with coefficients matrix $M$:

\begin{align}
    \varphi_{ij}(z) &= 
    \bmx{ 
    \mathbf{w}_{i,j,\pi(z,j)} & \cdots & \mathbf{w}_{i,j,\pi(z,j)+k} } M 
    \bmx{
    \theta_j(z)^k \\
    \vdots \\
    \theta_j(z)^2 \\
    \theta_j(z) \\
    1 \\
    },
    \label{eq:varphi_components}
\end{align}
where the matrix \(M\) for third-order uniform B-splines is

\begin{align}
M = \frac{1}{12} \bmx{
        -2 & 6 & -6 & 2 \\
         6 & -12 & 0 & 8\\
         -6 & 6 & 6 & 2 \\
        2 & 0 & 0 & 0
    }.
\end{align}
Additional $M$ matrices for higher-order B-splines as well as more detail on the non-recursive definition of B-splines can be found in \cite{christensen2024realtime}.

Note that in order to match the exact recursive definition of an open uniform B-spline, the $M$ matrices at the edge of the domain grid need to be recomputed from the recursive definition. In order to avoid this additional complexity, we use Equation \ref{eq:varphi_components} over the whole domain grid for each activation function. Our experiments give empirical evidence that this formulation achieves similar performance to the setup used in the original KAN paper (See Section \ref{results:feynman}).

\subsection{Adapting the Domain Grid with the Original KAN}
\label{sec:orig_kan}

Unlike MLPs that allow arbitrary inputs and return arbitrary outputs, the original Kolmogorov–Arnold Network (and B-splines in general) require a carefully defined grid interval that represents the domain of the control points of the B-splines. 
An important question to consider is how and when to stretch or prune the domain of the B-spline control points (weights). 

In the original KAN paper \cite{liu2024kan}, three hyperparameters are chosen to adapt the domain interval $[a, b]$ for each activation function. These hyperparameters control how often the grid is updated, when the grid updates start, and when they stop. Every $N$ steps (if it is before the start update step and after the stop update step) a manual grid update happens in the following way:


First, the algorithm feeds in a batch of data, sorts it along each data feature to give $x_{\text{sorted}}$, and evaluates the B-splines on the sorted data to give $y_{\text{eval}}$. Note that there is a separate domain grid for each data feature. After evaluating the sorted data, the algorithm picks different quantiles across each data feature to be the domain (grid) points for new sets of B-spline coefficients. These ``non-uniform" grids are blended with a grid with uniform spacing using a blending parameter $\epsilon$. Using the blended grid points and $y_{\text{eval}}$, the algorithm then finds a new set of coefficients (weights) with the $\texttt{curve2coef}$ function. Full details for this algorithm can be found in \cite{liu2024kan} and the associated codebase.

\section{Main Results}
\label{sec:main_results}

We now present our main results on the AdaptKAN architecture and histogram-based OOD detection method. We first describe AdaptKAN's structure and algorithm for auto-adapting the domain grid based on incoming data. We then outline our proposed algorithm for performing OOD detection using the histogram structures leveraged by AdaptKAN.

The AdaptKAN architecture consists of the KAN network described in Section \ref{sec:kan_overview} where each input feature in each layer is augmented with a histogram. The components of each of these histograms are as follows: 
\begin{itemize}
\item Each histogram (referenced as \texttt{hist} in Algorithm \ref{alg:histogram-adapt}) has a number of bins equal to the number of current grid intervals 
\item Each histogram has two additional bins, one below the minimum grid interval and one above the maximum grid interval, to keep track of out-of-domain data (referenced as \texttt{ood\_hist} in Algorithm \ref{alg:histogram-adapt}). These are represented by the black histogram bins in Figure \ref{fig:main_diagram}.
\item Each histogram has two variables that keep track of the minimum and maximum values seen during the entire training process (referenced as \texttt{ood\_a} and \texttt{ood\_b}, respectively). These values are outside the interval $[a, b]$ and are used to stretch the domain during training.
\end{itemize}
A single AdaptKAN layer has \(N_F N_B + 4\) extra parameters beyond the trainable KAN parameters described in Section \ref{sec:kan_overview}. An illustration of the architecture is given in Figure \ref{fig:main_diagram}.

Additionally, we introduce a few extra hyperparameters for the network that can stay fixed across various applications: a ``stretch" mode that indicates when the grid domain should stretch, an exponential moving average $\alpha$ rate which decides how much to weigh previous histograms and prune patience $p$ which effects how long we wait before we prune the grid domains. The possible ``stretch" modes are described in more detail in Appendix \ref{results:ablation} and EMA $\alpha$ and prune patience $p$ are utilized in Algorithm \ref{alg:histogram-adapt}.

\subsection{Auto-Adapting the Domain Grid with AdaptKAN}

\begin{figure*}[!t]
\centering
\includegraphics[width=\textwidth]{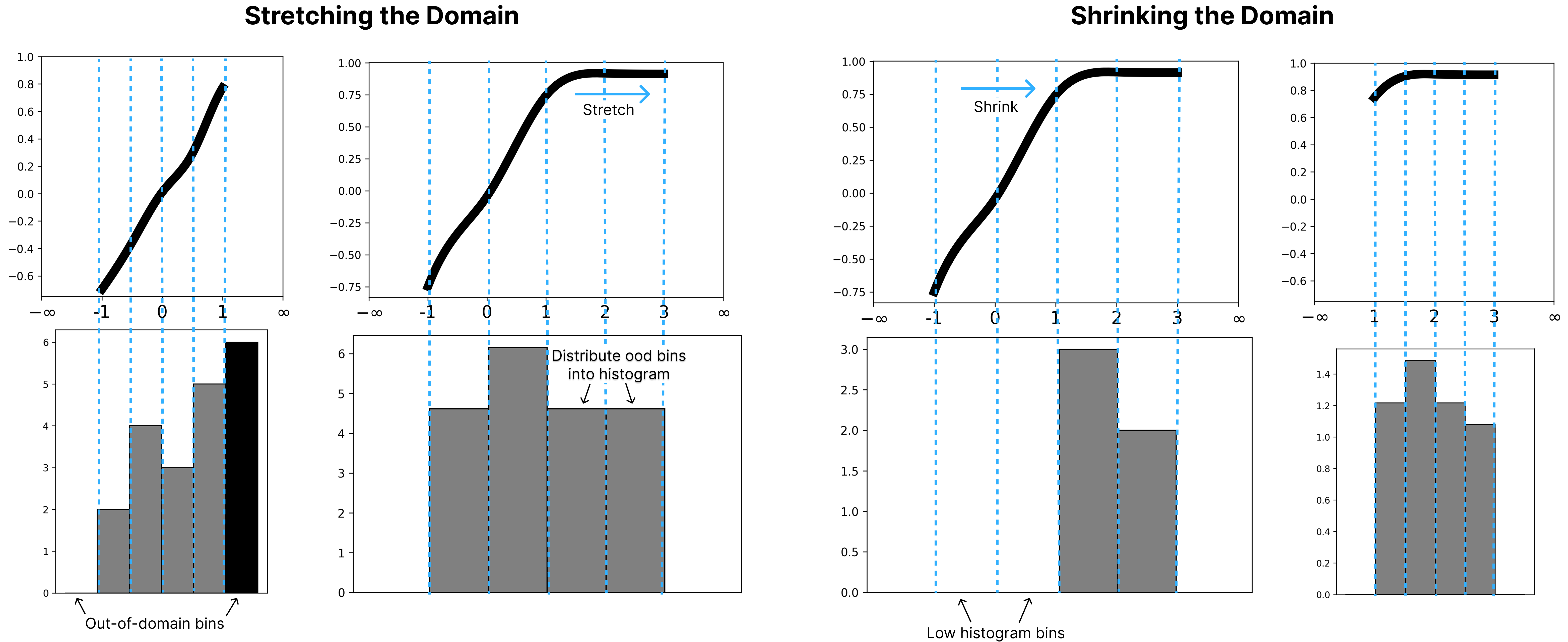}
\caption{How auto-adapting works: we look at out-of-domain bins (high values stretch the domain grid) and edge bins (close to empty bins shrinks the domain grid). We keep the number of weights and bins fixed during the training process and instead adjust the domain bounds $a$ and $b$. We refit the weights exactly using linear least squares after expansion/shrinking. Another, faster option, is approximately refitting the weights by interpolating over the control points (weights). We refit the histograms using linear interpolation, and we scale the histogram counts to equal the batch size.}
\label{fig:diagram}
\end{figure*}

In contrast to prior KAN architectures, AdaptKAN does not need to pick the hyperparameters for update frequency and the start/stop update steps. At a high level, our proposed approach performs two tasks. First, during training, the algorithm stores and updates an exponential rolling average of each of the input data distributions to each layer (referred to as \texttt{hist}). These marginal distributions for each feature are stored as histograms that can be utilized for other applications later on. For each input feature, the algorithm also keeps a rolling average of how much data is falling outside the domain grid (referred to as \texttt{ood\_hist}). This is a histogram with two bins only, a bin to track data falling below the domain grid and a bin to track falling above the domain grid. A detailed description of this can be seen in Algorithm \ref{alg:histogram-ema-update}. In this algorithm, we assume that the inputs $x_j$ are 1-dimensional, but this can easily be extended to higher dimensions by repeating the process for each dimension. Note that the \texttt{create\_histogram} function creates a histogram of the incoming data $X_{ID}$ over the interval $[a, b]$ with $n_{grid}$ bins with uniform bin widths.

\begin{algorithm}[ht]
\caption{Histogram Update with Exponential Moving Average (EMA)}  

\label{alg:histogram-ema-update}
\begin{algorithmic}[1]
\raggedright
\REQUIRE Batch~of~samples $\{x_j\}_{j=1}^B$, \\
\hspace*{1em} Domain bounds ~$a \in \mathbb{R}, b\in \mathbb{R}$, \\
\hspace*{1em} Number~of~bins~$n_{\mathrm{grid}}$, \\
\hspace*{1em} Previous~histogram$\mathrm{hist} \in \mathbb{R}^{n_{\mathrm{grid}}}$, \\
\hspace*{1em} EMA $\alpha \in [0,1]$
\ENSURE Updated~$\mathrm{hist}$, out-of-domain~histogram \\
\hspace*{1em} $\mathrm{ood\_hist}$ \\
\hspace*{1em} out-of-domain~values~$\mathrm{ood\_a}$, $\mathrm{ood\_b}$
\STATE $\mathrm{batch\_ood\_hist_1} \leftarrow \left| \left\{ j \in \{1,\ldots,B\} \;\middle|\; x_j < a \right\} \right|$
\STATE $\mathrm{batch\_ood\_hist_2} \leftarrow \left| \left\{ j \in \{1,\ldots,B\} \;\middle|\; x_j > b \right\} \right|$
\STATE $\mathrm{ood\_a} \leftarrow \min \left( \left\{ x_j \;\middle|\; x_j < a \right\}_{j=1}^B \cup \left\{\mathrm{ood\_a}\right\}\right)$
\STATE $\mathrm{ood\_b} \leftarrow \max \left( \left\{ x_j \;\middle|\; x_j > b \right\}_{j=1}^B \cup \left\{\mathrm{ood\_b}\right\}\right)$
\STATE $X_{ID} \leftarrow \left\{x_j \middle| x_j \geq a \land x_j \leq b \right\}_{j=1}^B$ 
\STATE $\mathrm{batch\_hist} \leftarrow \mathrm{create\_histogram}(X_{ID}; [a,b], n_{\mathrm{grid}})$
\STATE $\mathrm{hist} \leftarrow (1-\alpha) \cdot \mathrm{hist} + \alpha \cdot \mathrm{batch\_hist}$
\STATE $\mathrm{ood\_hist} \leftarrow (1-\alpha) \cdot \mathrm{ood\_hist} + \alpha \cdot \mathrm{batch\_ood\_hist}$
\STATE \textbf{return} $\mathrm{hist}$, $\mathrm{ood\_hist}$, $\mathrm{ood\_a}$, $\mathrm{ood\_b}$ \\
\end{algorithmic}
\vspace{0.5em}
\footnotesize\textit{Note: \cite{Claude} was used to generate latex code for this algorithm from project source code. This was checked and modified by the authors to ensure accuracy.}
\end{algorithm}

Second, at each training step our algorithm checks to see if the domain grid requires stretching or shrinking. Recall that the minimum and maximum values seen during training are tracked by the variables \texttt{ood\_a} and \texttt{ood\_b}, respectively. If enough input data is falling outside the domain 
the grid is stretched to cover the interval $[\texttt{ood\_a}, \texttt{ood\_b}]$. ``Enough" data is defined by a ``stretch" threshold, and possible options for this are detailed in Appendix \ref{results:ablation}. 
Otherwise, if the minimum or maximum in-domain histogram bins drop below a pre-defined ``shrink" threshold, we shrink the grid to exclude these near-empty bins. 

In either case where the domain is changed (stretching or shrinking), we refit the current weights to the new interval using linear least squares. Another faster option is to refit the weights directly using linear interpolation. This is done by finding the Greville abscissae \cite{farin2014curves} 
and linearly interpolating the weights to the new interval using the translated Greville abscissae as weight domain positions. We also refit the layer histograms via linear interpolation (using translated bin centers) over the new interval. If the domain was stretched, we zero out \texttt{ood\_hist} and absorb in any nonzero bin values into the corresponding layer histogram. When shrunk, nonzero bin values from \texttt{hist} are absorbed into \texttt{ood\_hist}. When refitting a layer histogram and \texttt{ood\_hist} to a new domain grid, we rescale by the sum of all the bins (referred to as the total histogram count) to equal the previous total histogram count (including \texttt{ood\_hist}). This is in order to maintain a consistent total count for each histogram throughout the training process. The auto-adapt method can be seen in more detail in Algorithm \ref{alg:histogram-adapt}. Note that in our experiments we used a slight variant of Algorithm \ref{alg:histogram-adapt} on line 12 (refilling the histogram bins); however this was later revised to more clearly align with what is outlined in this paper. The small revisions we made are outlined in the codebase associated with this paper.

\begin{algorithm}[ht]
\caption{Adaptive Histogram Domain Resizing} 
\label{alg:histogram-adapt}
\begin{algorithmic}[1]
\raggedright
\REQUIRE Current~domain~$[a,b]$, \\ 
\hspace*{1em} layer~histogram~$\mathrm{hist}$, \\
\hspace*{1em} OOD~histogram~$\mathrm{ood\_hist}$ \\
\hspace*{1em} OOD~values~$\mathrm{ood\_a}$, $\mathrm{ood\_b}$, prune~patience~$p$, \\
\hspace*{1em} EMA~rate~$\alpha$, network~weights~\texttt{coef}
\ENSURE Updated~domain~$[a,b]$, weights $\texttt{coef}$, layer histogram $\mathrm{hist}$, OOD histogram $\mathrm{ood\_hist}$
\STATE $\tau \leftarrow (1 - \alpha)^p \cdot \alpha$
\IF{$(\mathrm{ood\_hist_1} < \tau \wedge \mathrm{hist}_1 < \tau) \;\vee\; (\mathrm{ood\_hist_2} < \tau \wedge \mathrm{hist}_{n_{\mathrm{grid}}} < \tau)$}
  \STATE \textit{// shrink domain}
  \STATE $a \leftarrow$ first~bin~edge~where~$\mathrm{hist}_i > \tau$
  \STATE $b \leftarrow$ last~bin~edge~where~$\mathrm{hist}_i > \tau$
\ENDIF
\IF{$\mathrm{ood\_hist_1} > \max (\mathrm{hist}) \,\vee\, \mathrm{ood\_hist_2} > \max(\mathrm{hist})$}
  \STATE \textit{// stretch domain}
  \STATE $a \leftarrow \mathrm{ood\_a}$, $b \leftarrow \mathrm{ood\_b}$
\ENDIF
\STATE $\texttt{coef} \leftarrow$ Refit~on~$[a,b]$~using~linear~least~squares
  \STATE $\texttt{hist} \leftarrow$ Refit~on~$[a,b]$~using~linear~interpolation~filling~bins~with~\texttt{ood\_hist}~as~needed
  \STATE $\texttt{ood\_hist} \leftarrow$ Fill~bins~with~\texttt{hist}~where~needed~or~zero~out~bins
\STATE \textbf{return} $[a, b]$, $\texttt{coef}$, $\texttt{hist}$, $\texttt{ood\_hist}$
\end{algorithmic}
\vspace{0.5em}
\footnotesize\textit{Note: \cite{Claude} was used to generate latex code for this algorithm from project source code. This was checked and modified by the authors to ensure accuracy.}
\end{algorithm}

We choose a default value of $\tau_{shrink} = (1-\alpha)^p \cdot \alpha$ for our ``shrink" threshold. The reason for this choice can be illustrated by the following example involving a dataset containing outlier datapoints. Assume we see an outlier data point that only appears at the end of the first epoch but does not show up for the rest of the training cycle. Ideally, we would like our algorithm to prune the domain grid of the network so it can ``forget" the effects of that outlier point. When the data point is first input into the AdaptKAN network, it adds a value of $\alpha$ to the corresponding histogram bin for that outlier feature. If this datapoint does not show up for the in the second epoch, the algorithm will multiply that histogram bin by $(1-\alpha)$ a total of $p$ times, where $p$ is set to the number of batches per epoch. At the end of the second epoch, it will have a value of $(1-\alpha)^p \cdot \alpha$. Here we assume that no valid datapoint appears in the same histogram bin as the outlier, and this is the only outlier point. Selecting this value as the default pruning threshold results in the histogram bin containing the outlier being pruned at the end of the second epoch. Note that if there are N outlier points fed into the network falling into the same outlier bin (with no other datapoints in that bin), we would want a ``shrink" threshold of $\tau = N \cdot(1-\alpha)^p \cdot \alpha$ in order to shrink the domain after the end of the second epoch. We acknowledge that it may be difficult to predict in advance how many outlier points will be present in a given scenario. However, our experimental results demonstrate that this default $\tau_{shrink}$ threshold performs well on a variety of tasks. 

Another alternative to shrinking the domain with the fixed threshold is to shrink with a relative threshold. This is done by picking $\tau_{rel} = \max(\mathrm{hist}) \cdot \alpha$. This method did not perform as well in our experiments as $\tau_{shrink}$, so our algorithm opts $\tau_{shrink}$ threshold as the default. However, it is possible that this relative threshold method may be useful for other problem settings. We leave an investigation into this direction for future work.

Our method introduces new hyperparameters including the exponential moving average (EMA) weighting factor $\alpha$, the prune patience $p$, and ``stretch" thresholds. These hyperparameters can be set to sensible default values that can be used across different applications. For example, the EMA can stay the same across different applications, and $p$ is picked by default to be equal to the number of batches per epoch (or simply left at 1). The pruning patience tells us that we need to go though all the training data before we make a decision on pruning. We also have various ``stretch" modes that determine when to stretch the domain grids of the network. The effects of these parameters are studied in our experimental analyses (See Appendix \ref{results:ablation}). 

\subsection{Histogram Out-of-Distribution Detection}
\label{sec:ood}

Determining if data falls outside of a neural network's training distribution is important in situations where safety is a priority. OOD detection is a method to predict the likelihood that a previously unseen data point is from the same probability distribution of the original dataset that it was trained on. We now demonstrate that the layer histograms used for stretching and shrinking the domain grid can be used to form an OOD detection algorithm.

Consider a dataset of inputs $\mathcal{D}=\{x^i\}_{i=1}^n$ drawn i.i.d. from the joint distribution $\mathbf{P}$ over $\R^{N_F}$ with corresponding joint probability density function (PDF) $p(x_1,\dots,x_{N_F})$. Here $N_F$ is the number of features in our input data. We define the following events: 
\begin{equation}
\mathcal{E}_j^k=\{x \in \R^{N_F} | a_j + (k-1) d_j \leq x_j \leq a_j + k d_j\},
\end{equation}
where $a_j$, $b_j$ and $d_j$ are defined in Section \ref{sec:kan_overview} and $k \in \{1,\dots,\omega_j\}$. 
Assuming that no data from the $j$th feature fall outside the interval $[a_j, b_j]$, a histogram can be used to approximate $\mathbf{P}(\mathcal{E}_j^1) \dots \mathbf{P}(\mathcal{E}_j^\omega)$. This is done by counting the number times the input datapoints from the dataset \(\mathcal{D}\) fall into each of the $\omega_j$ evenly spaced bins from $a_j$ to $b_j$ and normalizing the bin values so that their sum is 1. Thus, an incoming datapoint with components that fall within low-value histogram bins can be interpreted as unlikely to be from the original input distribution and flagged as OOD data.
For the $j$th input feature, define the value \(P_j(x_j)\)
as the normalized bin value for \(x_j\), where
\begin{align}
P_j(x_j) = \frac{\texttt{histogram}_j[k^*(x_j)]}{\sum_{k=1}^{N_B} \texttt{histogram}_j[k]}.
\end{align}

Here $k^*(x_j)$ is the bin index (out of $N_B$ bins) that value $x_j$ falls into. Since $P_j(x_j) \approx \mathbf{P}\left( \mathcal{E}_j^{k^*(x_j)} \right)$, we can can interpret $\log (P_i(x_i))$ as an approximation of log probability of event $\mathcal{E}_j^{k^*(x_j)}$. These approximate log probabilities can be summed up over all $N_F$ features, giving us a total score for the input vector \(x\):

\begin{align}
\mathrm{score_{hist}}(x) = \frac{1}{N_F}\sum_{j=1}^{N_F} \log(P_j(x_j)).
\end{align}

A lower score indicates that the entire input \(x\) is more likely out-of-distribution, and a higher score indicates \(x\) is more likely to be from the original distribution that the network was trained on. During evaluation, the mean AUROC (Area Under the Receiving Operator Characteristic) score is calculated at varying thresholds to determine where to draw the line between OOD and in-distribution data. In practice, we pick a reasonable threshold based on in-distribution and OOD examples.

This method can be applied post-hoc to a trained AdaptKAN network. In other words, our approach does not require any extra training with an OOD-related loss function and/or OOD examples. This proposed histogram approach can also be combined with other post-hoc methods. For example, Maximum Softmax Probability (MSP) is an relatively effective post-hoc OOD detection method that involves taking the maximum over softmax outputs of the classification layer of a network \cite{hendrycks2016baseline}. The reasoning behind MSP is that the maximum softmax probability will be higher for in-distribution data than for OOD data, since OOD data is hypothesized to result in higher uncertainty and thus lower probability scores.  

It is possible to combine our proposed histogram method with MSP in the following manner: 

\begin{align}
\mathrm{score_{hist+msp}}(x) = \mathrm{score_{hist}}(x) + \lambda \log\left(\max_i \frac{e^{f(x)_i}}{\sum_{j=1}^K e^{f(x)_i}}\right)
\end{align}

Here, $f(x)$ is the logits output of the classification network and $\lambda$ is a hyperparameter that balances the weighting between the MSP and the histogram score.  

This proposed histogram approach strikes a balance between accuracy and scalability. The functions \(P_j(\cdot)\) can be interpreted as approximations of the marginal probability mass functions for a discretization of the joint PDF \(p(x_1,\ldots,x_{N_F})\), and so a complete characterization of the original distribution using our method is not possible. However, approximating the marginals rather than the joint PDF avoids the curse of dimensionality and allows our method to be applied to high-dimensional inputs with large numbers of features.
Experimental results show that our proposed method works well in practice and achieves state-of-the-art results on the OpenOOD v1.5 benchmark (see Section \ref{results:ood}). Future work will explore improvements and modifications to this approach, including replacing MSP with a method compatible with regression networks.


\section{Experimental Results}
\label{sec:experimental_results}

In this section, we evaluate AdaptKAN in several problem settings. Since one of the main benefits of the KAN architecture is symbolic formula representation \cite{yu2024kan}, we investigate how AdaptKAN compares to the original KAN when it comes to learning a variety of scientific equations, known as the Feynman equations (see Section \ref{results:feynman}). As an extension of these experiments, we analyze the effect of different AdaptKAN hyperparameters for the auto-adapt feature and compare the auto-adapt feature to the na\"ive manual adaptation (see Appendix \ref{results:ablation}). Additionally, we explore the application of larger AdaptKAN networks to image classification tasks via learning from high dimensional frozen features (see Section \ref{results:classification}). Then, we investigate the application of AdaptKAN to a robotics and controls task of learning a control Lyapunov function. We compare AdaptKAN's performance to that of an MLP using conformal prediction and test the effect of outliers on the proposed grid adaptation mechanism. Finally, 
we demonstrate experimentally that our proposed OOD detection method
achieves state-of-the-art OOD detection results as compared to prior methods (see Section \ref{results:ood}). Note that for the Feynman, image classification, and control Lyapunov experiments, we initialized the AdaptKAN network weights with a fixed seed value of zero, but we used different random seeds for the other elements of the experiments (such as data generation). 
\subsection{Feynman Equations}
\label{results:feynman}

We compare the performance of AdaptKAN (with the auto-adapt feature), MLPs and the original KAN on the Feynman dataset \cite{udrescu2019ai, udrescu2020ai} which is a collection of physics equations from Feynman's textbooks. The original paper introducing the KAN architecture \cite{liu2024kan} performs a variety of experiments comparing human constructed, pruned, and unpruned KANs against MLPs. Our experiments train the original KAN from scratch because the original paper did not provide the input ranges for the dimensionless formulas they used. As a result, our experimental procedures transform the data input ranges manually. This was done by picking ranges that avoided dividing by values close to zero, as well as capping largest and smallest values at $\pm$4 for some equations and a max of $2 \pi$ for others.

Similar to \cite{liu2024kan}, the original KAN architecture is trained with LBFGS initialized at $G=3$ grid intervals. Grid refinement (increasing the number of grid intervals for the same domain) happens every 200 steps to cover the schedule $G=\{3, 5, 10, 20, 50\}$. We use the default grid adaptation setup provided in the \texttt{pykan} repo and outlined in Section \ref{sec:orig_kan}. Tests are performed using 0, 1, or 2 prune rounds where the network is pruned using the sparsification technique described in \cite{liu2024kan}. We experiment with sparsification $\lambda = 10^{-2}, 10^{-3}$ or $10^{-4}$ and experiment pruning over 0, 1, or 2 rounds of 200 steps (before grid refinement). Like the original paper, the KAN has a fixed width of 5 and the depth is swept over values of 2, 3, 4, and 5. 

AdaptKAN training is performed using the Adam and AdamW optimizers. AdaptKAN networks are trained for 2000 steps each round. The network is refined from $G=3$ to $G=50$, covering the same grid schedule as the KAN. Similar to the KAN setup, we prune over 0, 1 and 2 rounds and use the same sparsification $\lambda$ values as the KAN. The number of AdaptKAN layers is swept over the values 2, 3, 4 and 5. We experiment with the following learning rate schedules: $lr=\{10^{-2}, 5\times10^{-3}, 10^{-3}, 5\times10^{-4}, 10^{-4}\}$ (with $lr=10^{-2}$ during the prune rounds) and $lr \in \{10^{-3}, 5\times10^{-4}, 10^{-4}, 5\times10^{-5}, 10^{-5}\}$ (with $lr=10^{-3}$ during the prune rounds). For the AdamW optimizer, we set a polynomial quadradic decay to 1/10th the learning rate at each point in our schedule (also using weight decay of $10^{-5}$ and Nesterov momentum). The stretch mode is set to ``half\_max'' (see Appendix \ref{results:ablation} for details on the different stretch modes), the value of EMA $\alpha$ is set to $10^{-3}$, the ``kan'' initialization is used, and activation initialization noise is set to 0.5. The ``kan'' initialization strategy is the same initialization strategy used in \cite{liu2024kan}, where a SiLU activation function applied to the inputs is added to the learned B-spline. Both spline and SiLU activation are multiplied by extra learned weights.

Finally, for the MLP we set up the experiments as follows: we sweep over various learning rate schedules with seven rounds of 2000 steps each using the Adam optimizer. We choose seven rounds of 2000 steps to have a fair comparison with AdaptKAN and KAN which can have a max of seven training rounds (with two rounds of pruning included). The learning rate schedules start with either $lr=10^{-2}$ or $10^{-3}$ and and we follow the same learning rate as AdaptKAN with the additional two extra rounds being $lr=\{5\times10^{-5}, 10^{-5}\}$ and $lr=\{5\times10^{-6}, 10^{-6}\}$ respectively. We sweep over depths in the set $\{2,5,10\}$ and widths in the set $\{5,10,20\}$. Activations are swept over $\text{ReLU}, \text{SiLU},$ and $\text{Tanh}$. We likewise sweep over the Adam and AdamW optimizers.

We record the lowest test Root Mean Squared Error (RSME) loss over all training rounds (measuring after each training round is finished) and five random seeds for each architecture configuration. Table \ref{table:feynman} shows the results of our experiments. Note that during training, issues arose with $\texttt{NaN}$ values being thrown for the original KAN architecture during training, typically when pruning was involved. When gathering the best results, these runs are excluded. However, the percentage of training runs where $\texttt{NaN}$s were thrown can be seen in Table \ref{table:feynman} as the ``Fail \%" metric across the different architectures.

\begin{table*}[!t]
\centering
\caption{Feynman Equations and Network Architectures}
\label{table:feynman}
\scriptsize
\setlength{\tabcolsep}{3pt}
\renewcommand{\arraystretch}{1.2}
\begin{tabular}{lp{3.8cm}ccccccccc}
\toprule
Feynman Eq. & Dimensionless Formula & \multicolumn{3}{c}{\# Parameters} & \multicolumn{3}{c}{Fail \%} & \multicolumn{3}{c}{Pairwise Comparisons} \\
\cmidrule(lr){3-5} \cmidrule(lr){6-8} \cmidrule(lr){9-11}
 & & \begin{tabular}{@{}c@{}}AdaptKAN\end{tabular} & \begin{tabular}{@{}c@{}}KAN\end{tabular} & \begin{tabular}{@{}c@{}}MLP\end{tabular} & \begin{tabular}{@{}c@{}}AdaptKAN\end{tabular} & \begin{tabular}{@{}c@{}}KAN\end{tabular} & \begin{tabular}{@{}c@{}}MLP\end{tabular} & \begin{tabular}{@{}c@{}}AdaptKAN\\vs KAN\end{tabular} & \begin{tabular}{@{}c@{}}AdaptKAN\\vs MLP\end{tabular} & \begin{tabular}{@{}c@{}}MLP\\vs KAN\end{tabular} \\
\midrule
I.6.2 & $\exp(-\theta^{2}/2\sigma^{2})/\sqrt{2\pi\sigma^{2}}$ & 350 & 910 & 3881 & 0.0 & 33.7 & 0.0 & Tie & AdaptKAN & KAN \\[0.5ex]
I.6.2b & $\exp\!\big(-(\theta-\theta_{1})^{2}/2\sigma^{2}\big)/\sqrt{2\pi\sigma^{2}}$ & 300 & 1805 & 3881 & 0.0 & 27.1 & 0.2 & AdaptKAN & Tie & Tie \\[0.5ex]
I.9.18 & $a\big/\big((b-1)^{2}+(c-d)^{2}+(e-f)^{2}\big)$ & 525 & 840 & 3881 & 0.0 & 46.4 & 0.2 & KAN & Tie & KAN \\[0.5ex]
I.12.11 & $1\big/(1+a\sin\theta)$ & 135 & 760 & 1781 & 0.1 & 7.7 & 0.2 & Tie & AdaptKAN & KAN \\[0.5ex]
I.13.12 & $a(1/b-1)$ & 176 & 171 & 1781 & 0.3 & 13.8 & 0.2 & Tie & Tie & Tie \\[0.5ex]
I.15.3x & $\dfrac{1-a}{\sqrt{1-b^{2}}}$ & 225 & 171 & 1781 & 0.0 & 11.0 & 0.2 & Tie & AdaptKAN & KAN \\[0.5ex]
I.16.6 & $\dfrac{a+b}{1+ab}$ & 275 & 760 & 1781 & 0.0 & 24.9 & 0.2 & AdaptKAN & AdaptKAN & KAN \\[0.5ex]
I.18.4 & $\dfrac{1+ab}{1+a}$ & 450 & 180 & 3881 & 0.0 & 20.4 & 0.2 & Tie & AdaptKAN & KAN \\[0.5ex]
I.26.2 & $\arcsin\!\big(n\sin\theta_{2}\big)$ & 600 & 1160 & 3881 & 0.1 & 11.6 & 0.2 & Tie & Tie & KAN \\[0.5ex]
I.27.6 & $\dfrac{1}{1+ab}$ & 2250 & 910 & 3881 & 0.0 & 29.8 & 0.2 & Tie & AdaptKAN & KAN \\[0.5ex]
I.29.16 & $p(1+a^{2}-2a\cos(\theta_{1}-\theta_{2}))$ & 1425 & 528 & 3881 & 0.0 & 22.7 & 0.2 & Tie & MLP & Tie \\[0.5ex]
I.30.3 & $\dfrac{\sin^{2}(n\theta/2)}{\sin^{2}(\theta/2)}$ & 975 & 658 & 3881 & 0.3 & 8.3 & 0.2 & KAN & Tie & Tie \\[0.5ex]
I.30.5 & $\arcsin\!\bigl(a/n\bigr)$ & 225 & 210 & 3881 & 0.0 & 23.2 & 0.2 & Tie & AdaptKAN & Tie \\[0.5ex]
I.37.4 & $1+a+2\sqrt{a}\cos\delta$ & 375 & 261 & 1781 & 0.3 & 10.5 & 0.2 & KAN & AdaptKAN & KAN \\[0.5ex]
I.40.1 & $n_{0}e^{-a}$ & 90 & 180 & 1781 & 0.0 & 22.7 & 0.2 & Tie & AdaptKAN & KAN \\[0.5ex]
I.44.4 & $nlna$ & 150 & 285 & 1781 & 0.0 & 18.8 & 0.2 & Tie & AdaptKAN & KAN \\[0.5ex]
I.50.26 & $\cos a + \alpha\cos^{2}a$ & 495 & 435 & 1781 & 0.0 & 8.3 & 0.2 & Tie & AdaptKAN & KAN \\[0.5ex]
II.2.42 & $(a-1)b$ & 175 & 114 & 1781 & 0.0 & 15.5 & 0.2 & Tie & AdaptKAN & KAN \\[0.5ex]
II.6.15a & $\dfrac{c}{\sqrt{a^{2}+b^{2}}}$ & 760 & 980 & 3881 & 0.0 & 44.8 & 0.2 & Tie & Tie & Tie \\[0.5ex]
II.11.17 & $n_{0}(1+a\cos\theta)$ & 180 & 1330 & 3881 & 0.4 & 8.8 & 0.2 & Tie & Tie & Tie \\[0.5ex]
II.11.27 & $\dfrac{n^{\alpha}}{1-n^{\alpha}/3}$ & 225 & 560 & 1781 & 0.1 & 15.5 & 0.2 & Tie & AdaptKAN & KAN \\[0.5ex]
II.35.18 & $\dfrac{n_{0}}{\exp(a)+\exp(-a)}$ & 375 & 285 & 1781 & 0.0 & 42.0 & 0.2 & Tie & AdaptKAN & KAN \\[0.5ex]
II.36.38 & $a+\alpha b$ & 160 & 152 & 1781 & 0.0 & 12.7 & 0.2 & Tie & Tie & KAN \\[0.5ex]
II.38.3 & $ab$ & 225 & 210 & 3881 & 0.0 & 26.5 & 0.2 & Tie & AdaptKAN & KAN \\[0.5ex]
III.9.52 & $a\,\dfrac{\sin^{2}\bigl((b-c)/2\bigr)}{((b-c)/2)^{2}}$ & 240 & 630 & 1781 & 0.0 & 40.9 & 0.2 & Tie & Tie & KAN \\[0.5ex]
III.10.19 & $1/\sqrt{1+a^{2}+b^{2}}$ & 575 & 42 & 3881 & 0.3 & 35.4 & 0.2 & Tie & AdaptKAN & KAN \\[0.5ex]
III.17.37 & $\beta\bigl(1+\alpha\cos\theta\bigr)$ & 2375 & 630 & 3881 & 0.1 & 19.9 & 0.2 & Tie & Tie & KAN \\[0.5ex]
\bottomrule
\end{tabular}
\\[6pt]
\begin{minipage}{\linewidth}
\footnotesize
\textit{Note:} Pairwise comparisons indicate statistically significant wins with Welch's two-tailed t-test. ``AdaptKAN", ``KAN", or ``MLP" indicates the significantly better architecture; ``Tie" indicates no significant difference at $p > 0.05$. Fail \% represents the percentage of runs that resulted in NaN values. \# Parameters is the smallest trainable parameter count observed across 5 random seeds for each best architecture. We note that the $0.2\%$ failures for the MLP happened with one particular configuration on a single seed: a depth of 2, width of 5, and ``tanh" activation. While it is not immediately clear why this configuration resulted in NaN values for most runs, we note that this network had far fewer parameters (51 total) than the top performing MLP configurations.
\end{minipage}
\end{table*}

The experimental results in Table 1 demonstrate that AdaptKAN matches the performance of the original KAN and outperforms the MLP with statistical significance. 
Additionally, AdaptKAN demonstrates significantly lower failure percentages than the KAN and lower failure percentages than the MLP. Finally, note that the number of trainable parameters for AdaptKAN and KAN are significantly lower on average than the number of trainable parameters for the MLP.

\subsection{Image Classification}
\label{results:classification}

To investigate AdaptKAN's performance with somewhat larger networks than used in the previous section, we perform image classification on frozen features for two popular datasets: CIFAR-10 and CIFAR-100 \cite{krizhevsky2009learning}. This was done as a quick way to test how AdaptKAN performs compared to medium sized ($\sim$1M parameter) MLPs in a simple classification task. One way to test image classification ability would have been to test the network on the raw features; however, since MLPs (and KANs) are not particularly suited for image classification we chose to treat each one as a ``classification head" trained on a ``frozen" pretrained model.
Three architectures are tested in these experiments: AdaptKAN, the original KAN architecture, and an MLP. These experiments use an alternative software implementation of the original KAN called EfficientKAN (used in \cite{canevaro2025advancing}), and as such we will refer to the KAN architecture as ``EfficientKAN" in this section.

\begin{figure*}[!t]
\centering
\includegraphics[width=\textwidth]{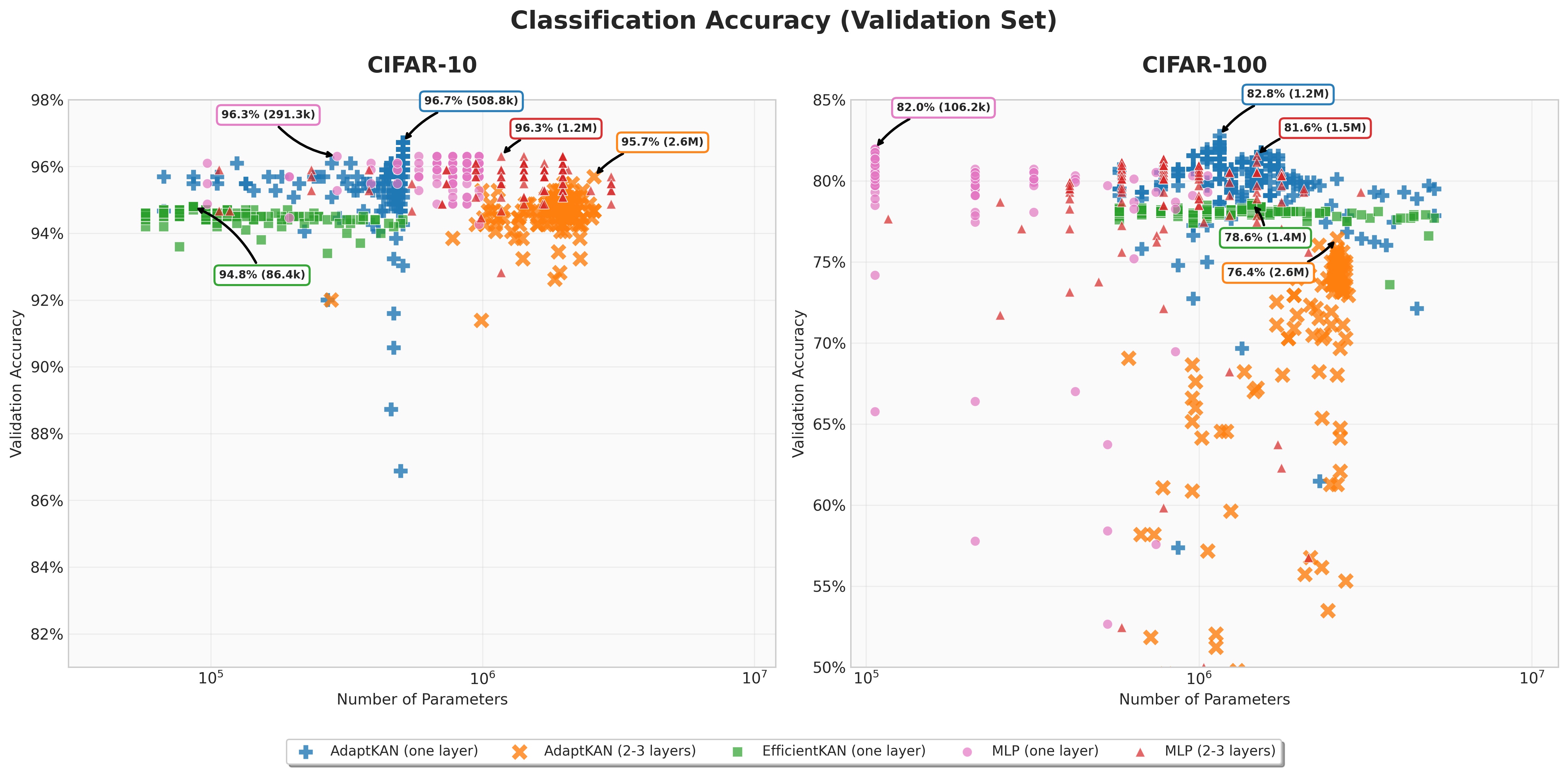}
\caption{We plot the validation accuracy results vs. the parameter count of various hyperparameter sweeps for AdaptKAN (one layer), Adaptkan (2-3 layers), EfficientKAN (one layer), MLP (one layer) and MLP (2-3 layers) on the CIFAR-10 and CIFAR-100 classification tasks. For CIFAR-10, AdaptKAN achieves a higher validation accuracy while achieving a lower number of parameters compared to the MLP (2-3 layers) best performing run. For CIFAR-100, AdaptKAN also achieves the highest accuracy, with fewer parameters than the best performing EfficientKAN and MLP (2-3 layers) model.}
\label{fig:class_acc}
\end{figure*}

In order to set up the image classification experiments, we use the training pipeline code from \cite{canevaro2025advancing}. Additionally, we use pretrained frozen features from ResNet18 models trained by \cite{zhang2023openood}. We use the $\texttt{TPESampler}$ from Optuna \cite{akiba2019optuna} to do hyperparameter sweeps of AdaptKAN, the MLP and EfficientKAN. 
For AdaptKAN and EfficientKAN, these experiments sweep over a variety of grid sizes. For all architectures we sweep over a variety of regularization $\lambda$ values, learning rate values, boolean flags on whether or not to normalize the input features, network width values, and network depth values. A comparison of Accuracy vs. Parameters can be seen in Figure \ref{fig:class_acc}. The implementation for EfficientKAN was not able to perform the $\texttt{update\_grid}$ method without throwing \texttt{NaN} values. As a result, these experiments only consider EfficientKAN architectures with one layer, which does not require a grid update step.
To facilitate direct comparisons, we divide our AdaptKAN and MLP results into those with a one-layer architecture and those with architectures having 2-3 layers.

These results show that AdaptKAN's classification accuracy performs at a level that exceeds or matches MLPs and EfficientKAN for networks with $\sim$1M parameters. The AdaptKAN (single layer) hyperparameter search for CIFAR-10 yielded a network with higher overall accuracy and a lower median number of parameters 
than the MLP architectures with 1 layer and 2-3 layers.
The median number of parameters over the Bayesian optimization hyperparameter search for AdaptKAN was 489.6k, the median number of parameters for the 1-layer MLP architectures was 873.9k, and the median number of parameters for MLP architectures with 2-3 layers was 1.97M. 
For CIFAR-100, AdaptKAN outperforms the MLP and EfficientKAN in overall accuracy while having a smaller median number of parameters (1.25M) compared to EfficientKAN (1.44M parameters). The higher parameter count compared to the MLP likely has to do with the larger output size of CIFAR-100 compared to CIFAR-10 (100 vs. 10 classes) and the associated extra parameters needed for KANs. 

\subsection{Learning a Control Lyapunov Function}
\label{results:dist_drift}

In this section, we compare the performances of the AdaptKAN and MLP architectures on the task of learning a control Lyapunov function (CLF). CLFs provide stability certificates for nonlinear dynamical systems \cite{sontag1989universal}. 
Our experiments in this section
consider the following dynamic system:
\begin{equation}
\dot{x} = f(x) + g(x) u,
\end{equation}
where
\begin{equation}
\label{eq:ccc_system}
f(x)=\begin{bmatrix} x_2^3 \\ -x_1^3 \end{bmatrix} \quad \text{and} \quad g(x) = \begin{bmatrix} 1 \\ 0 \end{bmatrix}
\end{equation}
In order to render the equilibrium point \([0, 0]^\intercal\) of the dynamics \eqref{eq:ccc_system} asymptotically stable, we seek to learn a neural CLF $V$ represented by, e.g., a KAN or MLP which must satisfy the following property:

\begin{equation}
\label{eq:nec_cond}
\pth{\frac{\partial V}{\partial x}g(x) = 0 \quad \land \quad x\neq 0} \implies \frac{\partial V}{\partial x}f(x)<0
\end{equation}

A positive definite function $V(x)$ that satisfies (\ref{eq:nec_cond}) in some neighborhood of the origin is a valid CLF. Given a CLF $V$, a stabilizing state feedback control is given by Sontag's formula \cite[Formula 5.56]{sontag2013mathematical}:

\begin{equation}
u(x) = \begin{cases}
-\frac{L_f V(x) + \sqrt{(L_f V(x))^2 + (L_g V(x))^4}}{L_g V(x)} & \text{if } L_g V(x) \neq 0 \\
0 & \text{if } L_g V(x) = 0
\end{cases}
\end{equation}

Here $L_f V \coloneq \frac{\partial V}{\partial x} f$ is the Lie derivative of $V$ along f and $L_g V \coloneq \frac{\partial V}{\partial x} g$ is the Lie derivative of $V$ along g. Additional mathematical background on CLFs is given in \cite[Sec. 9.7]{khalil2015nonlinear}.

A neural network $f_{\theta}$ (AdaptKAN or MLP) can be used to represent the Lyapunov candidate directly: 
\begin{align}
\label{eq:direct}
V(x) &= f_{\theta}(x),
\end{align}
where the inputs to the network are the state $x$ and the output is a single value. In order to get $\frac{\partial V}{\partial x}$, the \texttt{jax.grad} function is utilized. Note that since $V$ needs to be positive definite, this method requires the network to learn a positive definite output structure with $V(x_0) = 0$ through the loss functions. A positive definite output structure can also be enforced with: 
\begin{align}
\label{eq:enforce}
V(x) &= \frac{1}{2} f_{\theta}(x)^T f_{\theta}(x),
\end{align}
making the learning process easier. In this case, the output $f_{\theta}(x)$ of our network can be a vector of arbitrary length.

Loss functions for training neural CLFs are given in the prior work \cite{dawson2023safe},
but resulting performance on the system \eqref{eq:ccc_system} in our experiments was poor in initial experiments. We therefore use several custom loss functions to obtain better performance on this specific problem.
The terms included in our loss function are as follows: first, the term \(\mathcal{L}_{origin} = (V(x_0))^2\) from \cite{dawson2023safe} enforces the Lyapunov candidate to be zero at the point \(x_0\). Next, we use a custom ``bowl" loss term to encourage convexity:
\begin{align}
\mathcal{L}_{bowl} = \frac{1}{N}\sum_{i=1}^N &\max \left(0, k_1\left\| x_i \right\| - V(x_i)\right) \nonumber \\
&+ \max \left(0,V(x_i) - k_2\left\| x_i \right\|\right)
\end{align}
We set $k_1=0.001$ and $k_2=10$ throughout our experiments. Next, we define two loss terms to help enforce (\ref{eq:nec_cond}). They are:
\begin{align}
\mathcal{L}_{f} = \frac{1}{N}\sum_{i=1}^N & m_i \cdot\max \left(0,-L_f V_i\right) \nonumber \\
&+ (1-m_i)\cdot\max \left(0, L_fV_i \right),
\end{align}
where \(m_i\) represents the following mask based on the values of $L_gV$ :
\begin{align}
m_i = \begin{cases}
1 & \text{if } (L_gV)_i > \tau \\
0 & \text{otherwise}
\end{cases}
\end{align}

Choosing $\tau$ sets a threshold where $L_gV$ is considered ``small". Notice that in the areas $L_gV$ is small, we want $L_fV<0$. In our experiments we also found that enforcing $L_fV>0$ when $L_gV$ is large made a substantial difference. Investigating theoretical explanations for this behavior is left for future work.

The second loss term that helps enforce (\ref{eq:nec_cond}) is the following:
\begin{align}
\mathcal{L}_{g} = \frac{1}{N}\sum_{i=1}^N & (1-m_i) \cdot\max \left(0,\tau - |L_gV_i + \epsilon|\right) 
\end{align}
This loss term focuses on points where the magnitude of $L_gV$ is small, seeking to increase the magnitude until it reaches $\tau$. Here, $\epsilon = 10 ^{-8}$ is a small value added to avoid numerical stability issues.

Finally, we introduce a ``positivity" loss used in training scenarios where the network output has the possibility of becoming negative:
\begin{align}
\mathcal{L}_{pos} = \frac{1}{N}\sum_{i=1}^N \max \left(0,-V(x_i)\right) 
\end{align}

The final overall loss function used in these experiments is defined as a weighted sum of all the loss terms discussed previously:
\begin{align}
\mathcal{L} = \lambda_1 \mathcal{L}_{origin} + \lambda_2 \mathcal{L}_f + \lambda_3 \mathcal{L}_g + \lambda_4 \mathcal{L}_{bowl} + \lambda_5 \mathcal{L}_{pos}
\end{align}

In order to compare AdaptKAN and MLP, the values of $\lambda_2 \in \{0, 10^{-2}, 10^{-1}, 1, 10\}$, $\tau \in \{10^{-3}, 10^{-2}, 10^{-1}, 1, 2\}$, and learning rate $\in \{10^{-1}, 10^{-2}, 10^{-3}\}$ are swept over, while keeping $\lambda_1=10$, $\lambda_3=1$ and $\lambda_4=1$. First, experiments are conducted while enforcing positivity in the architecture as seen in Equation \ref{eq:enforce}, and in these cases the positivity loss is excluded. Network output sizes of 1 and 10 are used for both AdaptKAN and MLP. As a part of our experiments, depths of 3,4 and 5 for the MLP (with hidden size 64) and 1,2 and 3 for AdaptKAN (with hidden size 10) are swept over. Finally, the ``max'' stretch mode, an EMA $\alpha$ of 0.01, and the `linear' initialization strategy (which involves initializing the spline activations as noisy linear functions) are utilized along with an initialization noise of 0.1. We also experiment with using the direct output of the network as shown in Equation \ref{eq:direct}. In these training runs, $\lambda_5=1$ and the positivity loss with a network output size of 1 are used with the depths mentioned earlier. Each configuration is run over 5 random seeds. A total of 8000 training and 2000 test points are sampled uniformly from $x\in[-3, 3] \times [-3, 3]$.

To evaluate the CLFs on this task, we use techniques from conformal prediction \cite{shafer2008tutorial}. Briefly, conformal prediction seeks to construct a probabilistically valid prediction region $C : \mathbb{R}^K \to \mathbb{R}$ so that $\text{Prob}(R^{(0)} \leq C(R^{(1)}, \ldots, R^{(K)})) \geq 1 - \delta.$
Here, $R^{(0)}, R^{(1)}, \ldots, R^{(K)}$ are exchangeable random variables. Given a failure probability $\delta$, we can compute this upper bound as $C(R^{(1)}, \ldots, R^{(K)}) \coloneq \text{Quantile}_{1-\delta}(R^{(1)}, \ldots, R^{(K)}, \infty)$. Evaluating the \(\text{Quantile}_{1-\delta}\) function can be done by sorting $R^{(1)}, \ldots, R^{(K)}$ in non-decreasing order and computing $\text{Quantile}_{1-\delta}(R^{(1)}, \ldots, R^{(K)}, \infty) = R^{(p)}$ with $p\coloneq\ceil{(K+1)(1-\delta)}$. Conversely, given a fixed constant $C^*$ it is possible to find the corresponding $\delta^*(C^*)$ such that $\text{Prob}(R^{(0)} \leq C^*) \geq 1 - \delta^*(C^*)$. This is given by:
\begin{equation*}
\delta^*(C^*) = 1 - \frac{|\{R^{(i)}\leq C^* :i=1,\ldots,K\}|}{K+1}
\end{equation*}
In our case, the exchangeable random variable $R^{(i)}$ is defined to be $R^{(i)} \coloneq \left\|x_F^{(i)} - 0\right\|_2 = \left\|x_F^{(i)} \right\|_2$ where $x_F^{(i)}$ is the final position of our dynamics after 10 seconds from the \(i\)th starting point $x^{(i)}_S$ using the controller derived by the learned CLF and $0$ is the origin. 
In words, \(R^{(i)}\) represents the final distance of the system from the origin at the simulation time limit.
We use the Runge-Kutta 4 method for trajectory simulations and start at a uniformly distributed random point in our domain. We sample 200 points for each of the 5 seeds and pool the results together into a total of 1000 points for each hyperparameter configuration.

We calculate different confidence ($1-\delta$) percentages at fixed error values, and we calculate the error values associated with fixed confidence percentages. Results for both AdaptKAN and the MLP are compared against the following analytical control Lyapunov candidate:
\begin{equation}
\label{eq:analytical_clf}
V = \frac{1}{2} \left( x_1^2 + x_2^2 + \left(x_1 - x_2\right)^2 \right)
\end{equation} 
Appendix \ref{sec:proof} gives a proof that this candidate is a valid CLF for the dynamics in Equation \ref{eq:ccc_system}. 
Results comparing the AdaptKAN, MLP, and the Analytical CLF can be seen in Table \ref{tab:combined_comparison}. We include the top performing hyperparameter configuration for each spot in the table. This table considers the confidence levels associated with reaching a certain distance (or error bound $C$) from the origin. More specifically, it considers the distances $C=0.5, 0.25, 0.1$ and the confidence level associated with reaching that distance. We see that the trajectories resulting from the AdaptKAN CLF have higher confidence percentages for converging to within distances of $C=0.25$ and $C=0.1$ from the origin after 10 seconds of simulation as compared to the MLP or analytical CLFs. 
In other words, the statistical confidence that the AdaptKAN CLF causes the system to converge to within a distance of 0.1 from the origin is higher than the confidence for the MLP or analytical CLFs, and the statistical confidence that the AdaptKAN CLF causes the system to converge to within a distance of 0.25 from the origin is higher than the confidence for the MLP or analytical CLFs.

\begin{figure*}[!t]
\centering
\subfloat[]{\includegraphics[width=0.33\textwidth]{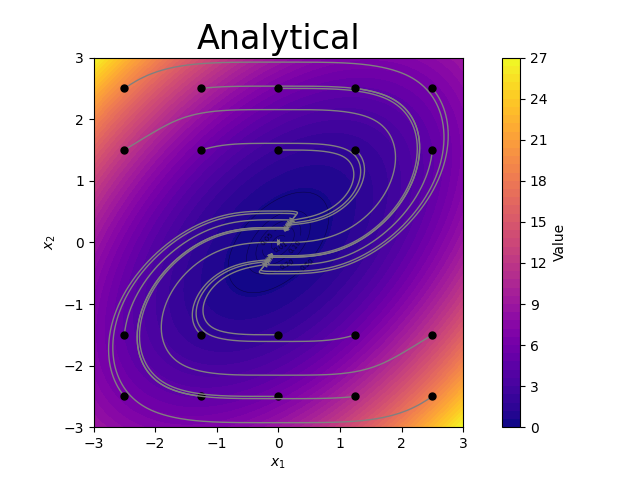}}%
\subfloat[]{\includegraphics[width=0.33\textwidth]{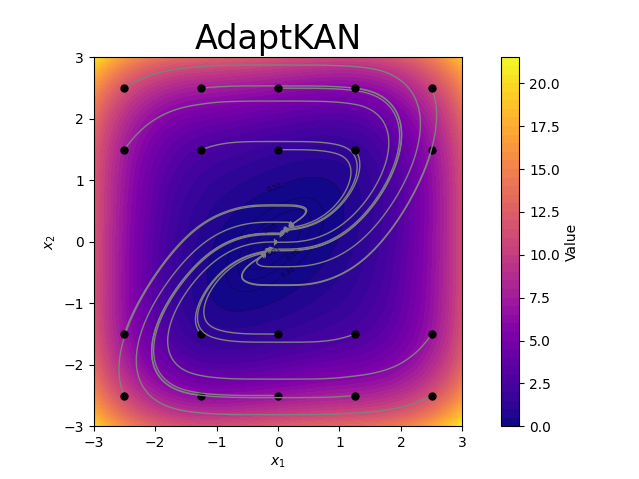}}%
\subfloat[]{\includegraphics[width=0.33\textwidth]{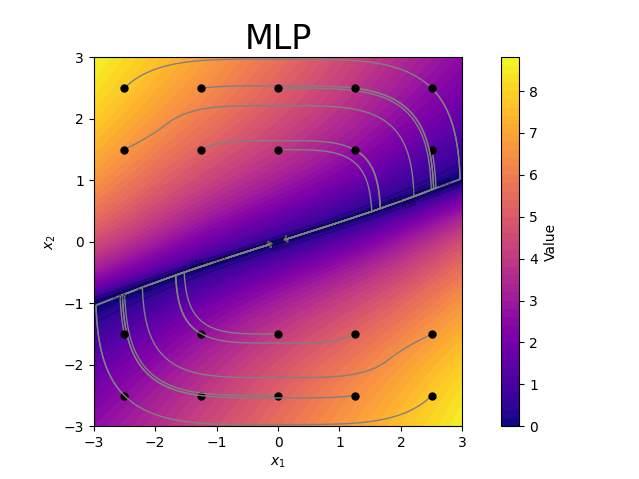}}
\caption{Lyapunov function contours and trajectories for the analytical solution (a), AdaptKAN (b), and MLP (c). In each of the plots, we simulate 20 trajectories (starting at a grid of points over the domain) using the controller derived from each CLF. In the background of the plots, we show the contours of the CLF values for each method, with each axis referring to an input feature. In these specific examples, 
AdaptKAN qualitatively seems to learn a CLF closer to the analytical solution. 
}
\label{fig:three_subfigures}
\end{figure*}

A qualitative depiction of the system trajectories and function shape for the three types of CLF tested is shown in Figure \ref{fig:three_subfigures}.

\begin{table*}[!t]
\centering
\caption{Comparison of Confidence Levels for different CLFs}
\label{tab:combined_comparison}

\begin{minipage}[t]{0.48\textwidth}
\centering
\captionsetup{labelformat=empty}
\label{tab:conformal_comparison}
\footnotesize
\setlength{\tabcolsep}{4pt}
\begin{tabular}{lcccc}
\toprule
\multirow{2}{*}{Function} & \multirow{2}{*}{Layers} & \multicolumn{3}{c}{Confidence Level \(1-\delta\) ($\uparrow$)} \\
\cmidrule(lr){3-5}
& & $C = 0.5$ & $C = 0.25$ & $C = 0.1$ \\
\midrule
Analytical & -- & \bf{0.999} & 0.252 & 0.129 \\
\midrule
AdaptKAN & 1 & \bf{0.999} & 0.674 & 0.169 \\
AdaptKAN & 2 & 0.894 & 0.753 & 0.433 \\
AdaptKAN & 3 & 0.929 & \bf{0.922} & \bf{0.785} \\
\midrule
MLP & 3 & 0.956 & 0.804 & 0.423 \\
MLP & 4 & \bf{0.999} & 0.715 & 0.393 \\
MLP & 5 & \bf{0.999} & 0.613 & 0.449 \\
\bottomrule
\end{tabular}
\end{minipage}%
\end{table*}

\subsubsection{Robustness to Noise}

To investigate the benefits of the proposed adaptation algorithm with the rolling exponential average histograms, we next present an experiment building upon the same control Lyapunov problem setup as mentioned previously.
This experiment investigates the model's robustness to noise added into training data at random points during training. The experiment procedures replace the training data with normally distributed noise in 1\% of the 1000 training epochs (a total of ten times). The noise is scaled by a value of 10 during five random epochs and scaled by $10^{-1}$ during another random five epochs. This represents real world scenarios where data can become corrupted or ``poisoned" occasionally. We compare the AdaptKAN ``adapt" function with a na\"ive manual adaptation method at different adaptation frequencies.

For the loss hyperparameters, we set $\lambda_2 = 0.1$ and $\tau=0.1$ for AdaptKAN, as these hyperparameters performed well in the experiments in the previous section. In addition, this experiment uses the ``half\_max'' method for the stretch mode, an EMA $\alpha$ of 0.001, the ``kan'' initialization strategy (described in Section \ref{results:feynman}), and an initialization noise of 0.5. A description of the stretch modes is given in Appendix \ref{results:ablation}. 
The na\"ive method performs a forced grid adaptation based on a single batch every $N$ training steps. We look at the lowest loss over all 5 random seeds and learning rates ($lr=\{10^{-1}, 10^{-2}, 10^{-3}\}$) for all configurations. Validation loss results at the end of training can be seen in Table \ref{tab:outliers}. The results demonstrate that AdaptKAN outperforms the na\"ive adaptation methods for different values of $N$ at three different depth values.

\begin{table}[!t]
\centering
\caption{Learning a CLF with Outliers}
\label{tab:outliers}
\footnotesize
\setlength{\tabcolsep}{4pt}
\begin{tabular}{lllc}
\toprule
Depth & Adapt Method & Learning Rate & Loss \\
\midrule
1 & Auto & 0.01 & $\boldsymbol{2.43 \times 10^{-2}}$ \\
1 & Manual (Every   1) & 0.01 & $8.45 \times 10^{-2}$ \\
1 & Manual (Every  10) & 0.01 & $8.28 \times 10^{-2}$ \\
1 & Manual (Every  50) & 0.01 & $8.39 \times 10^{-2}$ \\
1 & Manual (Every 100) & 0.01 & $8.65 \times 10^{-2}$ \\
1 & Manual (Every 200) & 0.01 & $8.09 \times 10^{-2}$ \\
1 & Manual (Every 500) & 0.01 & $8.09 \times 10^{-2}$ \\
\midrule
2 & Auto & 0.01 & $\boldsymbol{3.07 \times 10^{-2}}$ \\
2 & Manual (Every   1) & 0.01 & $4.76 \times 10^{-2}$ \\
2 & Manual (Every  10) & 0.001 & $6.58 \times 10^{-2}$ \\
2 & Manual (Every  50) & 0.001 & $6.00 \times 10^{-2}$ \\
2 & Manual (Every 100) & 0.1 & $5.72 \times 10^{-2}$ \\
2 & Manual (Every 200) & 0.001 & $6.54 \times 10^{-2}$ \\
2 & Manual (Every 500) & 0.01 & $4.71 \times 10^{-2}$ \\
\midrule
3 & Auto & 0.001 & $\boldsymbol{2.11 \times 10^{-2}}$ \\
3 & Manual (Every   1) & 0.01 & $4.70 \times 10^{-2}$ \\
3 & Manual (Every  10) & 0.01 & $3.98 \times 10^{-2}$ \\
3 & Manual (Every  50) & 0.01 & $4.33 \times 10^{-2}$ \\
3 & Manual (Every 100) & 0.01 & $4.21 \times 10^{-2}$ \\
3 & Manual (Every 200) & 0.01 & $4.41 \times 10^{-2}$ \\
3 & Manual (Every 500) & 0.01 & $4.62 \times 10^{-2}$ \\
\bottomrule
\end{tabular}
\end{table}

\subsection{Out-of-Distribution Detection}
\label{results:ood}

\newcommand{\result}[2]{#1{\scriptsize±#2}}
\begin{table*}[!t]
\centering
\caption{OOD Detection Performance Results on CIFAR-10 and CIFAR-100 Benchmarks}
\label{tab:ood_results}
\footnotesize
\setlength{\tabcolsep}{3.5pt}
\begin{tabular}{l*{9}{c}}  
\toprule
\multirow{2}{*}{Method} & \multicolumn{2}{c}{Near OOD} & \multicolumn{4}{c}{Far OOD} & \multirow{2}{*}{Avg Near} & \multirow{2}{*}{Avg Far} & \multirow{2}{*}{Avg Overall} \\
\cmidrule(lr){2-3} \cmidrule(lr){4-7}
& CIFAR & TIN & MNIST & SVHN & Textures & Places365 & & & \\
\midrule
\multicolumn{10}{c}{\textbf{CIFAR-10 Benchmark}} \\
\midrule
OpenMax & \result{86.91}{0.31} & \result{88.32}{0.28} & \result{90.50}{0.44} & \result{89.77}{0.45} & \result{89.58}{0.60} & \result{88.63}{0.28} & \result{87.62}{0.29} & \result{89.62}{0.19} & \result{88.95}{0.41} \\
ODIN & \result{82.18}{1.87} & \result{83.55}{1.84} & \result{95.24}{1.96} & \result{84.58}{0.77} & \result{86.94}{2.26} & \result{85.07}{1.24} & \result{82.87}{1.85} & \result{87.96}{0.61} & \result{86.26}{1.73}\\
MDS & \result{83.59}{2.27} & \result{84.81}{2.53} & \result{90.10}{2.41} & \result{91.18}{0.47} & \result{92.69}{1.06} & \result{84.90}{2.54} & \result{84.20}{2.40} & \result{89.72}{1.36} & \result{87.88}{2.05}\\
MDSEns & \result{61.29}{0.23} & \result{59.57}{0.53} & \result{\textbf{99.17}}{0.41} & \result{66.56}{0.58} & \result{77.40}{0.28} & \result{52.47}{0.15} & \result{60.43}{0.26} & \result{73.90}{0.27} & \result{69.41}{0.40}\\
RMDS & \result{88.83}{0.35} & \result{90.76}{0.27} & \result{93.22}{0.80} & \result{91.84}{0.26} & \result{92.23}{0.23} & \result{91.51}{0.11} & \result{89.80}{0.28} & \result{92.20}{0.21} & \result{91.40}{0.40} \\
Gram & \result{58.33}{4.49} & \result{58.98}{5.19} & \result{72.64}{2.34} & \result{91.52}{4.45} & \result{62.34}{8.27} & \result{60.44}{3.41} & \result{58.66}{4.83} & \result{71.73}{3.20} & \result{67.37}{5.04} \\
ReAct & \result{85.93}{0.83} & \result{88.29}{0.44} & \result{92.81}{3.03} & \result{89.12}{3.19} & \result{89.38}{1.49} & \result{90.35}{0.78} & \result{87.11}{0.61} & \result{90.42}{1.41} & \result{89.31}{1.96} \\
VIM & \result{87.75}{0.28} & \result{89.62}{0.33} & \result{94.76}{0.38} & \result{94.50}{0.48} & \result{95.15}{0.34} & \result{89.49}{0.39} & \result{88.68}{0.28} & \result{93.48}{0.24} & \result{91.88}{0.37} \\
KNN & \result{\textbf{89.73}}{0.14} & \result{\textbf{91.56}}{0.26} & \result{94.26}{0.38} & \result{92.67}{0.30} & \result{93.16}{0.24} & \result{\textbf{91.77}}{0.23} & \result{\textbf{90.64}}{0.20} & \result{92.96}{0.14} & \result{92.19}{0.27} \\
ASH & \result{74.11}{1.55} & \result{76.44}{0.61} & \result{83.16}{4.66} & \result{73.46}{6.41} & \result{77.45}{2.39} & \result{79.89}{3.69} & \result{75.27}{1.04} & \result{78.49}{2.58} & \result{77.42}{3.76} \\
SHE & \result{80.31}{0.69} & \result{82.76}{0.43} & \result{90.43}{4.76} & \result{86.38}{1.32} & \result{81.57}{1.21} & \result{82.89}{1.22} & \result{81.54}{0.51} & \result{85.32}{1.43} & \result{84.06}{2.16} \\
GEN & \result{87.21}{0.36} & \result{89.20}{0.25} & \result{93.83}{2.14} & \result{91.97}{0.66} & \result{90.14}{0.76} & \result{89.46}{0.65} & \result{88.20}{0.30} & \result{91.35}{0.69} & \result{90.30}{1.02} \\
NAC & \result{\textbf{89.83}}{0.29} & \result{\textbf{92.02}}{0.20} & \result{94.86}{1.37} & \result{96.06}{0.47} & \result{\textbf{95.64}}{0.45} & \result{\textbf{91.85}}{0.28} & \result{\textbf{90.93}}{0.23} & \result{94.60}{0.50} & \result{\textbf{93.37}}{0.64} \\
KAN & \result{\textbf{90.06}}{0.47} & \result{\textbf{91.92}}{0.52} & \result{97.86}{0.73} & \result{\textbf{97.39}}{0.42} & \result{\textbf{95.85}}{0.28} & \result{\textbf{91.64}}{0.91} & \result{\textbf{90.99}}{0.50} & \result{\textbf{95.69}}{0.22} & \result{\textbf{94.12}}{0.59} \\
MSP & \result{87.19}{0.33} & \result{88.87}{0.19} & \result{92.63}{1.57} & \result{91.46}{0.40} & \result{89.89}{0.71} & \result{88.92}{0.47} & \result{88.03}{0.25} & \result{90.73}{0.43} & \result{89.83}{0.76} \\

\textbf{Hist}  & \result{87.58}{0.68} & \result{90.28}{0.54} & \result{\textbf{98.61}}{0.87} & \result{\textbf{98.89}}{0.36} & \result{\textbf{95.51}}{0.38} & \result{87.46}{0.41} & \result{88.93}{0.58} & \result{\textbf{95.12}}{0.24} & \result{93.06}{0.57} \\
\textbf{Hist + MSP} & \result{87.80}{0.66} & \result{90.46}{0.54} & \result{\textbf{98.56}}{0.70} & \result{\textbf{98.69}}{0.39} & \result{95.14}{0.40} & \result{88.38}{0.60} & \result{89.13}{0.57} & \result{\textbf{95.19}}{0.20} & \result{\textbf{93.17}}{0.56} \\
\midrule
\multicolumn{10}{c}{\textbf{CIFAR-100 Benchmark}} \\
\midrule
OpenMax & \result{74.38}{0.37} & \result{78.44}{0.14} & \result{76.01}{1.39} & \result{82.07}{1.53} & \result{80.56}{0.09} & \result{79.29}{0.40} & \result{76.41}{0.25} & \result{79.48}{0.41} & \result{78.46}{0.88} \\
ODIN & \result{78.18}{0.14} & \result{81.63}{0.08} & \result{83.79}{1.31} & \result{74.54}{0.76} & \result{79.33}{1.08} & \result{79.45}{0.26} & \result{79.90}{0.11} & \result{79.28}{0.21} & \result{79.49}{0.77} \\
MDS & \result{55.87}{0.22} & \result{61.50}{0.28} & \result{67.47}{0.81} & \result{70.68}{6.40} & \result{76.26}{0.69} & \result{63.15}{0.49} & \result{58.69}{0.09} & \result{69.39}{1.39} & \result{65.82}{2.66} \\
MDSEns & \result{43.85}{0.31} & \result{48.78}{0.19} & \result{\textbf{98.21}}{0.78} & \result{53.76}{1.63} & \result{69.75}{1.14} & \result{42.27}{0.73} & \result{46.31}{0.24} & \result{66.00}{0.69} & \result{59.44}{0.93} \\
RMDS & \result{77.75}{0.19} & \result{82.55}{0.02} & \result{79.74}{2.49} & \result{84.89}{1.10} & \result{83.65}{0.51} & \result{\textbf{83.40}}{0.46} & \result{80.15}{0.11} & \result{82.92}{0.42} & \result{82.00}{1.15} \\
Gram & \result{49.41}{0.58} & \result{53.91}{1.58} & \result{80.71}{4.15} & \result{\textbf{95.55}}{0.60} & \result{70.79}{1.32} & \result{46.38}{1.21} & \result{51.66}{0.77} & \result{73.36}{1.08} & \result{66.12}{1.98} \\
ReAct & \result{\textbf{78.65}}{0.05} & \result{\textbf{82.88}}{0.08} & \result{78.37}{1.59} & \result{83.01}{0.97} & \result{80.15}{0.46} & \result{\textbf{80.03}}{0.11} & \result{\textbf{80.77}}{0.05} & \result{80.39}{0.49} & \result{80.52}{0.79} \\
VIM & \result{72.21}{0.41} & \result{77.76}{0.16} & \result{81.89}{1.02} & \result{83.14}{3.71} & \result{85.91}{0.78} & \result{75.85}{0.37} & \result{74.98}{0.13} & \result{81.70}{0.62} & \result{79.46}{1.62} \\
KNN & \result{77.02}{0.25} & \result{\textbf{83.34}}{0.16} & \result{82.36}{1.52} & \result{84.15}{1.09} & \result{83.66}{0.83} & \result{79.43}{0.47} & \result{80.18}{0.15} & \result{82.40}{0.17} & \result{81.66}{0.87} \\
ASH & \result{76.48}{0.30} & \result{79.92}{0.20} & \result{77.23}{0.46} & \result{85.60}{1.40} & \result{80.72}{0.70} & \result{78.76}{0.16} & \result{78.20}{0.15} & \result{80.58}{0.66} & \result{79.79}{0.69} \\
SHE & \result{78.15}{0.03} & \result{79.74}{0.36} & \result{76.76}{1.07} & \result{80.97}{3.98} & \result{73.64}{1.28} & \result{76.30}{0.51} & \result{78.95}{0.18} & \result{76.92}{1.16} & \result{77.59}{1.78} \\
GEN & \result{\textbf{79.38}}{0.04} & \result{\textbf{83.25}}{0.13} & \result{78.29}{2.05} & \result{81.41}{1.50} & \result{78.74}{0.81} & \result{\textbf{80.28}}{0.27} & \result{\textbf{81.31}}{0.08} & \result{79.68}{0.75} & \result{80.23}{1.10} \\
NAC & \result{72.02}{0.69} & \result{79.86}{0.23} & \result{93.26}{1.34} & \result{92.60}{1.14} & \result{\textbf{89.36}}{0.54} & \result{73.06}{0.63} & \result{75.94}{0.41} & \result{\textbf{87.07}}{0.30} & \result{\textbf{83.36}}{0.84} \\
KAN & \result{72.97}{0.17} & \result{81.37}{0.22} & \result{92.29}{1.85} & \result{87.16}{4.46} & \result{\textbf{89.43}}{0.39} & \result{77.42}{0.35} & \result{77.17}{0.17} & \result{86.57}{0.70} & \result{\textbf{83.44}}{1.99} \\
MSP & \result{\textbf{78.47}}{0.07} & \result{82.07}{0.17} & \result{76.09}{1.86} & \result{78.42}{0.89} & \result{77.32}{0.70} & \result{79.23}{0.29} & \result{\textbf{80.27}}{0.11} & \result{77.76}{0.44} & \result{78.60}{0.90} \\
\textbf{Hist} & \result{63.62}{1.36} & \result{72.94}{2.00} & \result{\textbf{97.45}}{0.60} & \result{\textbf{96.89}}{1.03} & \result{87.45}{0.58} & \result{64.73}{2.47} & \result{68.28}{1.68} & \result{\textbf{86.63}}{0.36} & \result{80.51}{1.51} \\
\textbf{Hist + MSP} & \result{70.96}{0.62} & \result{79.72}{0.78} & \result{\textbf{94.88}}{0.66} & \result{\textbf{96.15}}{1.09} & \result{\textbf{88.22}}{0.39} & \result{73.38}{1.52} & \result{75.34}{0.69} & \result{\textbf{88.16}}{0.20} & \result{\textbf{83.89}}{0.92} \\
\bottomrule
\end{tabular}
\end{table*}

We test the OOD method described in Section \ref{sec:ood} on the Open OOD v1.5 CIFAR10 and CIFAR100 benchmarks \cite{zhang2023openood}. For each benchmark, three pre-trained ResNet18 models are provided. For each pre-trained network, each OOD detection method is applied post-hoc on the features from various layers of the network. The mean AUROC (Area Under the Receiving Operator Characteristic) score is calculated by varying different possible thresholds of where to draw the line between OOD and in-distribution data.

Each benchmark is divided into Near OOD detection and Far OOD detection. Near OOD detection looks at datasets that are similar to the in-distribution data the ResNet backbone was trained on. For CIFAR-10, these datasets include CIFAR-100 and the TIN dataset. For CIFAR-100, these include CIFAR-10 and the TIN dataset. Far OOD detection looks at datasets that have substantially different semantics compared to the in-distribution data. For CIFAR-10 and CIFAR-100, these include the MNIST, SVHN, Textures, and Places365 datasets. 

In Table \ref{tab:ood_results}, we evaluate two novel post-hoc methods: one with the histograms only (created from the features of the pretrained backbone) and one with the histograms combined with MSP of the pretrained backbone. These novel methods are detailed in Section \ref{sec:ood}. We use the same pipeline code as the KAN OOD method \cite{canevaro2025advancing} to evaluate these methods, and we include the other scores provided in their paper. For the ``Hist" method, we use 200 histogram bins and for the ``Hist + MSP" method we use 50 histogram bins and $\lambda=10^{-1}$. The value of $\lambda=10^{-1}$ was verified to be a good choice via a hyperparameter search on validation data. 

We bold the top three best performing methods in Table \ref{tab:ood_results}. For both benchmarks, our ``Hist + MSP" method is in the top three post-hoc methods for Average Far OOD detection and Average Overall OOD detection. The ``Hist + MSP" method yields the best post-hoc results for the CIFAR-100 benchmark in the Average Far OOD and Average Overall OOD categories. We remain competative with Average Near OOD as well. We outperform and remain competitive with the KAN method, which involves an additional training process with significantly more parameters compared to our method. 

While we did not use the entire AdaptKAN architecture in these results, the process of doing OOD detection with AdaptKAN is straightforward. The layer histograms in AdaptKAN may effectively be treated as a built-in OOD detectors. It is also possible to combine the ``Hist" score with the MSP of the output of AdaptKAN for classification tasks. 
We note that while it is possible to build feature histograms for arbitrary architectures, it is not immediately clear how to select hyperparameters such as the upper and lower bounds of the features (handled by the \texttt{ood\_counts} in AdaptKAN) as well as the number of histogram bins (handled by the grid size of AdaptKAN). 
Future work will investigate the application of such feature histograms to other architectures.

\section{Conclusion}
\label{sec:conclusion}


This paper introduced a novel variant of the Kolmogorov-Arnold Network architecture, called AdaptKAN, that leverages a histogram method to adapt grid domains based on incoming data. The proposed method eliminates the need to select the timing for domain grid adjustment, simplifying and streamlining the KAN training process.
In addition, this paper introduced a novel OOD detection method based on the histogram architecture used by AdaptKAN. 
Experimental results demonstrate that the AdaptKAN architecture matches or exceeds the performance of prior architectures on several tasks involving learning symbolic equations, image classification, and learning control Lyapunov functions. 
The proposed OOD detection method achieves state-of-the-art results on the OpenOOD v1.5 benchmark.
Future work will focus on broadening the application of these methods to problems in scientific machine learning, robotics, and TinyML on low-SWaP devices.


{\appendices

\section{Feynman Hyperparameter Sweep Analysis}
\label{results:ablation}

In this section, we explore the impact a variety of different parameters have on the overall training results for the Feynman equations (see the Section \ref{results:feynman} for details on the Feynman equations). We set up the hyperparameter sweep experiments in the following way. We use the Adam optimizer with learning rate schedule $lr=\{10^{-2}, 5\times10^{-3}, 10^{-3}, 5\times10^{-4}, 10^{-4}\}$ (2000 steps for each learning rate), no prune rounds, a prune patience of 1, a network depth 2 or 3 layers, and a layer width of 5. We then sweep over the following configurations:

\subsubsection{Exact Weight Refitting} During the grid refinement and adapt mechanism, the weights have to be refit to the new grid domain. We explore either exactly refitting the weights via linear least squares or doing a rough refitting of the control points via linear interpolation, using the Greville abscissae \cite{farin2014curves} as the domain of the control points. This ``rough" refitting is much faster for larger AdaptKAN networks.

\subsubsection{Stretch Modes} We explore a variety of methods that tell us when to stretch the grid domain. These include ``max'' (stretch if our out-of-distribution bins exceed the max histogram bin), ``half\_max'' (stretch if our ood bins exceed half the max histogram bin), ``mean'' (stretch if our ood bins exceed the mean of the histogram bins), and ``edge'' (stretch if our ood bins exceed either of the edge histogram bins). Note that we do not expand/stretch one side of the domain after a trigger; we actually stretch both sides each time. We found that this method of stretching both sides at the same time performed similarly to the method of stretching a single side at a time on some of the tests we ran.

\subsubsection{EMA $\alpha$'s} We look at the effect of the exponential moving average $\alpha$ rate, which decides what percentage of past histograms should contribute to the updated histograms each step in the training process.

\subsubsection{Manual Adapting} We observe what happens when we replace AdaptKAN's ``adapt" method with a na\"ive grid adaptation method that forces the grid domain to adjust based on a single batch every N epochs. This is similar to the algorithm outlined in Section \ref{sec:orig_kan}, except that our grid interval sizes are all uniform. 

Table \ref{tab:ablation1} shows the results of the hyperparameter sweep. We see that the highest performing configuration for the Feynman dataset is the ``mean" stretch mode, an EMA $\alpha$ of 1.0, and the approximate refit method mentioned earlier. We also compare directly to the manual adaptation method in Table \ref{tab:ablation2}. The auto method outperforms all manual methods for the approximate refit scenario, and ties the manual adaptation method in the exact refit scenario.

\begin{table}[!t]
\centering
\caption{Feynman Loss Hyperparameter Sweep}
\label{tab:ablation1}
\footnotesize
\setlength{\tabcolsep}{4pt}
\begin{tabular}{lcccc}
\toprule
Stretch Mode & EMA $\alpha$ & Loss (Approx Refit) & Loss (Exact Refit) \\
\midrule
edge & 0.001 & $8.90 \times 10^{-4}$ & $8.52 \times 10^{-4}$ \\
edge & 0.010 & $1.27 \times 10^{-2}$ & $6.06 \times 10^{-3}$ \\
edge & 0.100 & $4.06 \times 10^{-2}$ & $1.38 \times 10^{-2}$ \\
edge & 0.500 & $5.05 \times 10^{-2}$ & $1.43 \times 10^{-2}$ \\
edge & 0.750 & $3.64 \times 10^{-2}$ & $8.60 \times 10^{-3}$ \\
edge & 0.900 & $3.31 \times 10^{-2}$ & $5.95 \times 10^{-3}$ \\
edge & 1.000 & $3.83 \times 10^{-4}$ & $5.41 \times 10^{-4}$ \\
half\_max & 0.001 & $6.63 \times 10^{-4}$ & $7.40 \times 10^{-4}$ \\
half\_max & 0.010 & $1.32 \times 10^{-3}$ & $1.10 \times 10^{-3}$ \\
half\_max & 0.100 & $1.83 \times 10^{-3}$ & $2.08 \times 10^{-3}$ \\
half\_max & 0.500 & $1.21 \times 10^{-3}$ & $1.63 \times 10^{-3}$ \\
half\_max & 0.750 & $1.19 \times 10^{-3}$ & $1.23 \times 10^{-3}$ \\
half\_max & 0.900 & $1.25 \times 10^{-3}$ & $1.43 \times 10^{-3}$ \\
half\_max & 1.000 & $4.17 \times 10^{-4}$ & $5.30 \times 10^{-4}$ \\
max & 0.001 & $7.62 \times 10^{-4}$ & $7.70 \times 10^{-4}$ \\
max & 0.010 & $1.02 \times 10^{-3}$ & $1.20 \times 10^{-3}$ \\
max & 0.100 & $1.52 \times 10^{-3}$ & $1.71 \times 10^{-3}$ \\
max & 0.500 & $1.16 \times 10^{-3}$ & $1.50 \times 10^{-3}$ \\
max & 0.750 & $1.32 \times 10^{-3}$ & $1.53 \times 10^{-3}$ \\
max & 0.900 & $1.23 \times 10^{-3}$ & $1.56 \times 10^{-3}$ \\
max & 1.000 & $3.86 \times 10^{-4}$ & $5.33 \times 10^{-4}$ \\
mean & 0.001 & $6.60 \times 10^{-4}$ & $7.35 \times 10^{-4}$ \\
mean & 0.010 & $1.35 \times 10^{-3}$ & $1.05 \times 10^{-3}$ \\
mean & 0.100 & $1.95 \times 10^{-3}$ & $1.99 \times 10^{-3}$ \\
mean & 0.500 & $1.07 \times 10^{-3}$ & $1.49 \times 10^{-3}$ \\
mean & 0.750 & $1.13 \times 10^{-3}$ & $1.28 \times 10^{-3}$ \\
mean & 0.900 & $1.19 \times 10^{-3}$ & $1.43 \times 10^{-3}$ \\
mean & 1.000 & $\boldsymbol{3.26 \times 10^{-4}}$ & $\boldsymbol{5.17 \times 10^{-4}}$ \\
\bottomrule
\end{tabular}
\end{table}

\begin{table}[!t]
\centering
\caption{Feynman Loss Auto vs Manual Adapt}
\label{tab:ablation2}
\footnotesize
\setlength{\tabcolsep}{4pt}
\begin{tabular}{lcc}
\toprule
Adapt Method & Loss (Approx Refit) & Loss (Exact Refit) \\
\midrule
Manual (Every 1) & $4.99 \times 10^{-4}$ & $3.85 \times 10^{-4}$  \\
Manual (Every 10) & $5.46 \times 10^{-4}$ & $3.24 \times 10^{-4}$  \\
Manual (Every 50) & $5.23 \times 10^{-4}$ & $\boldsymbol{2.85 \times 10^{-4}}$  \\
Manual (Every 100) & $4.39 \times 10^{-4}$ & $2.97 \times 10^{-4}$  \\
Manual (Every 200) & $3.96 \times 10^{-4}$ & $3.03 \times 10^{-4}$  \\
Manual (Every 500) & $3.06 \times 10^{-4}$ & $2.99 \times 10^{-4}$  \\
Auto\textsuperscript{*} & $\boldsymbol{2.07 \times 10^{-4}}$ & $\boldsymbol{2.85 \times 10^{-4}}$  \\
\bottomrule
\end{tabular}
\vspace{2pt}
\begin{flushleft}
\footnotesize
*Note: This uses a manual adapt after the first epoch (which was not used in the results for Table \ref{tab:ablation1}), an EMA $\alpha=1.0$, and the `edge' stretch mode. This is different than a manual adapt every epoch because we adapt only when the prune and stretch mechanisms trigger. For both Tables \ref{tab:ablation1} and \ref{tab:ablation2} we look at the median loss over all runs.
\end{flushleft}
\end{table}

\section{Proof of Stability for the Analytical CLF}
\label{sec:proof}

Consider the dynamics in Equation \ref{eq:ccc_system} and the following Lyapunov candidate from \eqref{eq:analytical_clf}: 
\begin{equation}
V = \frac{1}{2} \left( x_1^2 + x_2^2 + \left(x_1 - x_2\right)^2 \right)
\end{equation}
Observe that \(V\) is positive definite and radially unbounded.
We demonstrate that the candidate \(V\) satisfies the property in \eqref{eq:nec_cond} as follows. 
To determine when \(\pdrv{V}{x}g(x) = 0\),
observe that
\begin{align*}
    &\pdrv{V}{x}g(x) = \bmx{2x_1-x_2 \\ 2x_2 - x_1}^T \bmx{1 \\ 0} 
    = 2x_1 - x_2.
\end{align*}
Therefore \(\pdrv{V}{x}g(x) = 0\) when $x_2 = 2x_1$. Note that under this constraint, $x_1 \neq 0$ if and only if $x_2 \neq 0$. Making the substitution $x_2 = 2x_1$ and assuming $x\neq0$,
\begin{align*}
    &\pdrv{V}{x}f(x) = \bmx{2x_1-x_2 \\ 2x_2 - x_1}^T \bmx{x_2^3 \\ -x_1^3} \\
    &\hspace{3.5em}= \bmx{2x_1 - (2x_1) \\ 2(2x_1) - x_1}^T \bmx{(2x_1)^3 \\ -x_1^3} \\
    &\hspace{3.5em}= \bmx{0 \\ 3x_1}^T \bmx{8x_1^3 \\ -x_1^3} \\
    &\hspace{3.5em}= -3x_1^4 < 0
\end{align*}
By these arguments, \(\pdrv{V}{x} g(x) = 0 \implies \pdrv{V}{x} f(x) < 0\) for all \(x \neq 0\).
Thus, the Lyapunov candidate from \eqref{eq:analytical_clf} is a valid CLF.



}

\section*{Acknowledgments}
We would like to thank the Brigham Young University Office of Research and Computing for the use of their computing resources to conduct the experiments in this paper. AI assistance \cite{Claude} was used to generate latex code for Algorithms \ref{alg:histogram-ema-update} and \ref{alg:histogram-adapt} from project source code. This was checked and modified by the authors to ensure accuracy. AI assistance \cite{Claude,ChatGPT,Gemini} was also used to edit and generate portions of the project source code.

\section*{Availability of Data and Materials}
The source code to our project is available at: \url{https://github.com/byu-magicc/adaptkan}.

\bibliographystyle{IEEEtran}
\bibliography{references}

\newpage

 




\vfill

\end{document}